\documentclass[10pt,twocolumn,letterpaper]{article}

\usepackage[pagenumbers]{cvpr} 

\definecolor{cvprblue}{rgb}{0.21,0.49,0.74}
\usepackage[pagebackref,breaklinks,colorlinks,allcolors=cvprblue]{hyperref}
\usepackage{array}
\usepackage{multirow}
\usepackage{pifont}
\usepackage{arydshln}
\usepackage{makecell}

\def\paperID{8343} 
\def\confName{CVPR}
\def\confYear{2025}

\title{Neuro-3D: Towards 3D Visual Decoding from EEG Signals}

\author{%
	Zhanqiang Guo$^{1, 2*}$, Jiamin Wu$^{1, 3*}$, Yonghao  Song$^{2}$, Jiahui Bu$^{5}$, Weijian Mai$^{1, 6}$, \\ Qihao Zheng$^{1}$,  Wanli Ouyang$^{1, 3\dag}$, Chunfeng Song$^{1, 4\dag}$ \\[0.5em]
	$^{1}$Shanghai Artificial Intelligence Laboratory, $^{2}$Tsinghua University, \\$^{3}$The Chinese University of Hong Kong, $^{4}$Shanghai Innovation Institute,  \\$^{5}$Shanghai Jiao Tong University, $^{6}$South China University of Technology \\[0.3em] 
}

\begin{document}
	\maketitle
	\renewcommand{\thefootnote}{}
	\footnotetext{$^*$ Equal contribution. $^\dag $ Corresponding authors: \texttt{\{songchunfeng, ouyangwanli\}@pjlab.org.cn}}
	\renewcommand{\thefootnote}{\arabic{footnote}}
	
	\begin{abstract} 
	Human's perception of the visual world is shaped by the stereo processing of 3D information.
	Understanding how the brain perceives and processes 3D visual stimuli in the real world has been a longstanding endeavor in neuroscience. 
	Towards this goal, we introduce a new neuroscience task: decoding 3D visual perception from EEG signals, a neuroimaging technique that enables real-time monitoring of neural dynamics enriched with complex visual cues.
	To provide the essential benchmark, we first present EEG-3D, a pioneering dataset featuring multimodal analysis data and extensive EEG recordings from 12 subjects viewing 72 categories of 3D objects rendered in both videos and images.
	Furthermore, we propose Neuro-3D, a 3D visual decoding framework based on EEG signals. This framework adaptively integrates EEG features derived from static and dynamic stimuli to learn complementary and robust neural representations, which are subsequently utilized to recover both the shape and color of 3D objects through the proposed diffusion-based colored point cloud decoder.
	To the best of our knowledge, we are the first to explore EEG-based 3D visual decoding. Experiments indicate that Neuro-3D not only reconstructs colored 3D objects with high fidelity,  but also learns effective neural representations that enable insightful brain region analysis. The code and dataset are available at \url{https://github.com/gzq17/neuro-3D}.
	
\end{abstract}    
	\vspace{-0.3cm}
\section{Introduction}
\label{sec:intro}

\textit{“The brain is wider than the sky.”} — Emily Dickinson

The endeavor to comprehend how the human brain perceives the visual world has long been a central focus of cognitive neuroscience
~\cite{hendee1997perception,grill2004human}. 
As we navigate through the environment, 
our perception of the three-dimensional world is shaped by both fine details and the diverse perspectives from which we observe them. This stereo experience of color, depth, and spatial relationships forms complex neural activity in the brain's cortex.
Unraveling how the brain processes 3D perception remains an appealing challenge in neuroscience. Recently, electroencephalography (EEG), a non-invasive neuroimaging technique favored for its safety and ethical suitability, has been widely adopted in 2D visual decoding ~\cite{yi2024learning,chen2024eegformer,chen2023seeing,jiang2024large,luo2024brain,lahner2024modeling} to generate static visual stimuli.
With the aid of EEG and generative techniques, an intriguing question arises: \textit{\textbf{can we directly reconstruct the original 3D visual stimuli from dynamic brain activity?}}

\begin{figure}[t]
	\centering
	\includegraphics[width=0.8\columnwidth]{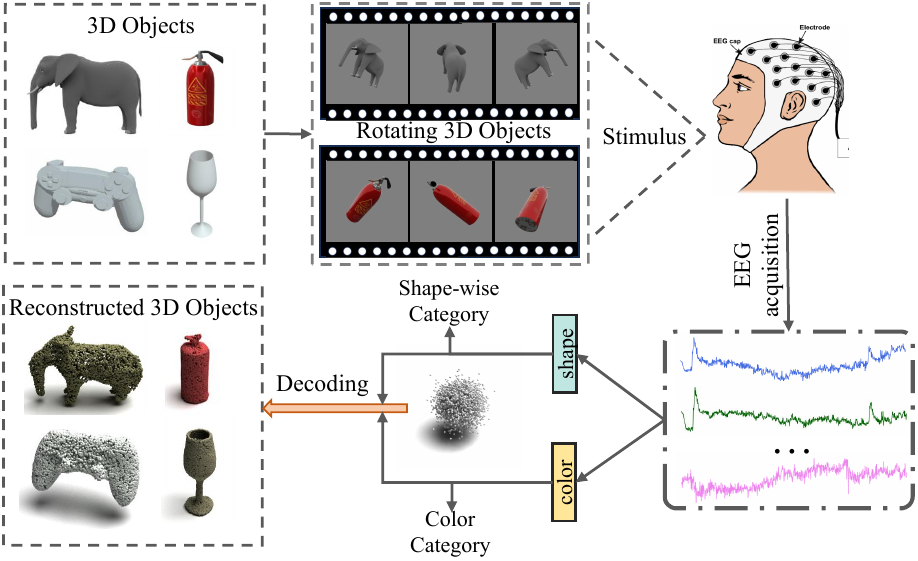} 
	\vspace{-0.1cm}
	\caption{Illustration of brain activity acquisition and colored 3D objects reconstruction from EEG signal.}
	\label{intro_img}
	\vspace{-0.5cm}
\end{figure}

To address this question, in this paper, we explore a new task, \textbf{3D visual decoding from EEG signals}, shedding light on the brain mechanisms for perceiving natural 3D objects in the real world.  To be specific, this task aims to reconstruct 3D objects from the EEG signals in the form of colored point clouds, as shown in Fig.~\ref{intro_img}.
The task involves not only extracting semantic features but also capturing intricate visual cues, \textit{e.g.}, color, shape, and structural information, underlying in dynamic neural signals, all of which are essential for a thorough understanding of 3D visuals.
In observing the surrounding world, humans 
form 3D perception through shifting views of objects in continuous movement overtime.
EEG provides an effective means of tracking neural dynamics in this evolving perceptual process for the 3D decoding task, owing to its high temporal resolution with millisecond precision~\cite{engel1994fmri,gifford2022large}. This property
distinguishes it from other neuroimaging techniques like fMRI with high spatial resolution but extremely low temporal resolution of few seconds~\cite{engel1994fmri}. 
Furthermore, as EEG offers the advantages of cost-effectiveness and portability, EEG-based 3D visual decoding research could be employed in real-time applications such as in clinical scenarios~\cite{metzger2023high,moses2021neuroprosthesis}.

However, when delving into this task, two critical challenges need to be addressed.
(1) \textbf{Limited data availability}: 
Currently, there is no publicly available dataset that provides paired EEG signals and 3D stimulus data. 
(2)~\textbf{Complexity of neural representation}:
The neural representations are inherently complex~\cite{hebb2005organization}. 
This complexity is amplified by low signal-to-noise ratio of non-invasive neuroimaging techniques, making it challenging to learn robust neural representation and recover complex 3D visual cues from brain signals.
Thus, how to construct a robust 3D visual decoding framework is a critical issue.

To address the first challenge, we develop a new EEG dataset, named \textbf{EEG-3D} dataset, comprising paired EEG signals collected from 12 participants while watching 72 categories of 3D objects. To create diverse 3D stimuli, we select a subset of common objects from the Objaverse dataset \cite{deitke2023objaverse,xu2023pointllm}. Previous works \cite{gao2023mind,sargent2023zeronvs} have revealed that 360-degree rotating videos effectively represent 3D objects. Thus, we capture rotational videos of colored 3D objects to serve as  visual stimuli, as shown in Fig.~\ref{intro_img}.
Compared to existing datasets \cite{horikawa2017generic,chang2019bold5000,wen2018neural,gao2023mind,kavasidis2017brain2image,gifford2022large,grootswagers2022human,allen2022massive}, EEG-3D dataset offers several distinctive features: 
(1)~\textbf{Comprehensive EEG signals in diverse states}. 
In addition to EEG signals from video stimuli, our dataset includes signals from static images and resting-state activity, providing diverse neural responses and insights into brain perception mechanisms across dynamic and static scenes.
(2)~\textbf{Multimodal analysis data with high-quality annotations}. The dataset comprises high-resolution videos, static images, text captions, and corresponding 3D objects with geometry and color details, supporting a wide range of visual decoding and analyzing tasks.
Building upon the EEG-3D dataset, 
we introduce an EEG-based 3D visual decoding framework, termed as \textbf{Neuro-3D}, to reconstruct 3D visual cues from complex neural signals. 
We first propose a Dynamic-Static EEG-Fusion Encoder to extract robust and discriminative EEG features against noises. 
Given EEG recordings evoked from dynamic and static stimuli, we design an attention based neural aggregator to adaptively fuse different EEG signals, exploiting their complementary characteristics to extract robust neural representation.
Subsequently, to recover 3D perception from EEG embedding, we propose a Colored Point Cloud Decoder, with the first stage generating the shape and the second stage assigning colors to the generated point clouds. To enhance precision in the generation process, we further decouple the EEG embedding into distinct geometry and appearance components, enabling targeted conditioning of shape and color generation.
To learn discriminative and semantically meaningful EEG features,
we align them with visual features of observed videos through contrastive learning \cite{wang2020understanding}. 
Finally, utilizing the aligned geometry feature as condition, a 3D diffusion model is applied to generate the point cloud of the 3D object, which is then combined with appearance EEG feature for color prediction. 
Our main contributions can be summarized as follows:

\begin{itemize}
	\item We are the first to explore the task of 3D visual decoding from EEG signals, which serves as a critical step for advancing neuroscience research into the brain’s 3D perceptual mechanism.
	\item We present \textbf{EEG-3D}, a pioneering dataset accompanied by both multimodal analysis data and comprehensive EEG recordings from 12 subjects watching 72 categories of 3D objects.
	This dataset fills a crucial gap in 3D-stimulus neural data for the computer vision and neuroscience communities.
	\item We propose \textbf{Neuro-3D}, a 3D visual decoding framework based on EEG signals. A diffusion-based colored point cloud decoder is proposed to recover both shape and color characteristics of 3D objects from adaptively fused EEG features captured under static and dynamic 3D stimuli. 
	\item The experiments indicate that Neuro-3D not only reconstructs colored 3D objects with high fidelity,  but also learns effective neural representation that enables insightful brain region analysis. 
\end{itemize}

	\section{Related Work}
\label{sec:related_work}

\subsection{2D Visual Decoding from Brain Activity}

Visual decoding from brain activity \cite{lahner2024modeling,yi2024learning,chen2023seeing,chen2024eegformer,chen2024cinematic,zhang2022neural} has gained substantial attention in computer vision and neuroscience, emerging as an effective technique for understanding and analyzing human visual perception mechanisms.
Early approaches predominantly utilized Convolutional Neural Networks (CNNs) and Generative Adversarial Networks (GANs) to model brain activity signals and interpret visual information~\cite{horikawa2017generic,shen2019deep,beliy2019voxels,lin2022mind,ozcelik2022reconstruction,huang2021fmri}. 
Recently, the utilization of newly-emerged diffusion models \cite{ho2020denoising,nichol2021improved} and vision-language models \cite{li2023blip,zhu2023minigpt} has advanced visual generation from various neural signals including fMRI \cite{chen2023seeing,takagi2023high,scotti2024reconstructing,sun2024contrast,scotti2024mindeye2}
and EEG \cite{song2023decoding,bai2023dreamdiffusion,li2024visual,singh2023eeg2image}. 
These methods typically perform contrastive alignment \cite{wang2020understanding} between 
neural signal embeddings and image or text features derived from the pre-trained CLIP model~\cite{radford2021learning}. Subsequently, the aligned neural embeddings are sent into diffusion model to conditionally reconstruct images that correspond to the visually-evoked brain activity. 
Apart from static images,
research has begun to extend these approaches to the reconstruction of video information from fMRI data, further advancing the field \cite{chen2024cinematic,sun2024neurocine,wen2018neural,wang2022reconstructing,kupershmidt2022penny}.
Though impressive, these methods, limited to 2D visual perception, fall short of capturing the full depth of human 3D perceptual experience in real-world environments. Our method attempts to expand the scope of brain visual decoding to three dimensions by reconstructing 3D objects from real-time EEG signals.

\subsection{3D Reconstruction from fMRI}

Reconstructing 3D objects from brain signals holds significant potential for advancing both brain analysis applications and our understanding of the brain’s visual system. To achieve this goal, several works~\cite{gao2023mind,fMRI-3D} have made initial strides in 3D object reconstruction from fMRI, yielding promising results in interpreting 3D spatial structures. 
Mind-3D~\cite{gao2023mind} proposes the first dataset of paired fMRI and 3D shape data and develops a diffusion-based framework to decode 3D shape from fMRI signals. A subsequent work, fMRI-3D~\cite{fMRI-3D}, expanded the dataset to include a broader range of categories across five subjects. 

However, there are several limitations in previous task setup, which prevent it from simulating real-time, natural 3D perception scenarios. First, fMRI equipment is unportable, expensive, and difficult to operate, potentially hindering its application in BCIs (Brain-Computer-Interface). Besides its high acquisition cost, fMRI is limited by its inherently low temporal resolution, which hinders real-time responsiveness to dynamic stimuli. Second, existing brain 3D reconstruction methods focus exclusively on reconstructing 3D shape of objects, neglecting color and appearance information that are crucial in real-world perception.
To address these challenges, 
we introduce a 3D visual decoding framework based on EEG signals, along with 
a new dataset for paired EEG and colored objects. 
To the best of our knowledge, this is the first work to interpret 3D objects from EEG signals, offering a comprehensive dataset, benchmarks, and decoding framework.

\vspace{-1mm}
\subsection{Diffusion Models}

Diffusion model has recently emerged as a powerful generative framework known for high-quality image synthesis capabilities. Inspired by non-equilibrium thermodynamics, the diffusion models are formulated as Markov chains. The model first progressively corrupts the target data distribution by adding noise until it conforms to a standard Gaussian distribution, and subsequently generates samples by predicting and reversing the noise process through network learning \cite{ho2020denoising,nichol2021improved}. The diffusion model, along with its variants, has been extensively applied to tasks such as image generation \cite{rombach2022high,kim2022diffusionclip,saharia2022photorealistic,ramesh2021zero} and image editing \cite{yu2024accelerating,zhang2023adding}.

Building on 2D image generation, \cite{luo2021diffusion} and \cite{zhou20213d} extended pixel-based approaches to 3D coordinates, enabling the generation of point clouds. This has spurred further research into 3D generation \cite{ren2024tiger,vahdat2022lion}, text-to-3D reconstruction \cite{poole2022dreamfusion,nichol2022point}, and 2D-to-3D generation \cite{melas2023pc2,wu2023sketch,long2024wonder3d}, demonstrating their capability to capture intricate spatial structures and textures of 3D objects.
In our study, we extend the 3D diffusion model to brain activity analysis, reconstructing colored 3D objects from EEG signals.
	\section{EEG-3D Dataset}
\label{dataset}

In this section, we introduce the detailed procedures for building the EEG-3D dataset.

\subsection{Participants}

We recruited $12$ healthy adult participants ($5$ males, $7$ females; mean age: $21.08$ years) for the study. All participants have normal or corrected-to-normal vision. Informed written consent was obtained from all individuals after a detailed explanation of the experimental procedures. Participants received monetary compensation for their involvement. The study protocol was reviewed and approved by the Ethics Review Committee.

\begin{figure}[t]
	\centering
	\includegraphics[width=0.73\columnwidth]{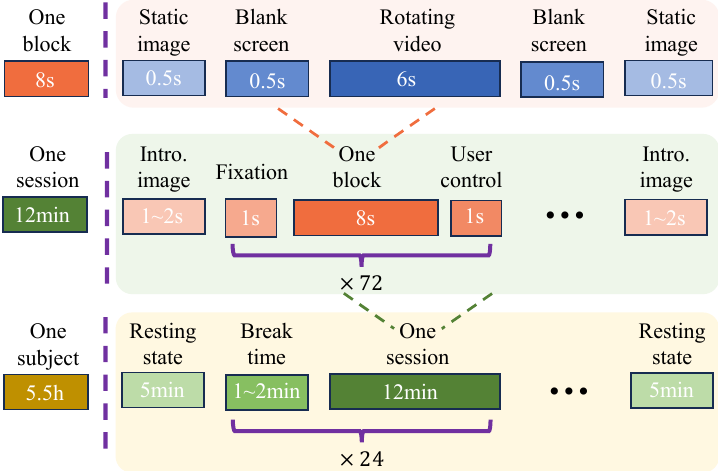} 
	\caption{The data collection process for one participant.}
	\label{data_acquisition}
	\vspace{-0.4cm}
\end{figure}

\subsection{Stimuli}

The stimuli employed in this study were derived from the Objaverse dataset \cite{deitke2023objaverse,xu2023pointllm}, which offers an extensive collection of common 3D object models. We selected 72 categories with different shapes, each containing 10 objects accompanied by text captions. For each category, 8 objects were randomly allocated to the training set, while the remaining 2 were reserved for the test set. 
Additionally, we assigned objects with color type labels, dividing them into six categories according to their main color style.
To generate the visual stimuli, we followed the procedure in Zero-123 \cite{liu2023zero} to use Blender to simulate a camera that captured 360-degree views of each object through incremental rotations, yielding 180 high-resolution images (1024 $\times$ 1024 pixels) per object. The objects were tilted at an optimal angle to provide comprehensive perspectives.

Rotating 3D object videos offer multi-perspective views, capturing the overall appearance of 3D objects. However, the prolonged duration of such videos, coupled with factors such as eye movements, blink artifacts, task load and lack of focus, often leads to EEG signals with a lower signal-to-noise ratio. In contrast, static image stimuli provide single-perspective but more stable information, which can serve to complement the dynamic EEG signals by mitigating their noise impact. Therefore, we collected EEG signals for both dynamic video and static image stimulus. The stimulus presentation paradigm is shown in Fig. \ref{data_acquisition}.
Specifically, the multi-view images were compiled into a 6-second video at 30 Hz. Each object stimulus block consisted of a 8-second sequence of events: a 0.5-second static image stimulus at the beginning and end, a 6-second rotating video, and a brief blank screen transition between each segment. 
During each experimental session, a 3D object was randomly selected from each category with a 1-second fixation cross between object blocks to direct participants' attention. Participants manually initiated each new object presentation. 
Training set objects had $2$ measurement repetitions, while test set objects had $4$, resulting in totaling $24$ sessions. 
Participants took 2-3 minute breaks between sessions.
Following established protocols \cite{gifford2022large}, 5-minute resting-state data were recorded at the start and end of all sessions to support further analysis. Each participant's total experiment time was approximately 5.5 hours, divided into two acquisitions.
\vspace{-0.3cm}
\subsection{Data Acquisition and Preprocessing}
\vspace{-0.1cm}
During the experiment, images and videos were presented on a screen with a resolution of 1920 × 1080 pixels. Participants were seated approximately 95 cm from the screen, ensuring that the stimuli occupied a visual angle of approximately 8.4 degrees to optimize perceptual clarity. EEG data were recorded using a 64-channel EASYCAP equipped with active silver chloride electrodes, adhering to the international 10-10 system for electrode placement. Data acquisition was conducted at a sampling rate of 1000 Hz. 
Data preprocessing was performed using MNE \cite{gramfort2013meg}, and more details are shown in \textbf{Supplementary Material}.

\begin{table}[t]
	\centering
	\renewcommand\arraystretch{1.0}
	\scriptsize
	\resizebox{1.0\columnwidth}{!}{
		\begin{tabular}{ p{52pt}<{\centering}  p{12pt}<{\centering}  p{12pt}<{\centering}  p{12pt}<{\centering}  p{12pt}<{\centering}  p{12pt}<{\centering}  p{12pt}<{\centering}  p{12pt}<{\centering}  p{12pt}<{\centering}}
			\hline
			\multirow{2}*{\textbf{Dataset}}  & \multicolumn{3}{c}{\multirow{1}{*}{\textbf{Brain Activity}}}
			& \multicolumn{5}{c}{\multirow{1}{*}{\textbf{\textbf{Analysis Data}}}} \\ 
			\cmidrule(l{2pt}r{2pt}){2-4} \cmidrule(l{2pt}r{2pt}){5-9}
			& \multirow{1}{*}{\textbf{Re}} & \multirow{1}{*}{\textbf{St}} & \multirow{1}{*}{\textbf{Dy}} & \multirow{1}{*}{\textbf{Img}} & \multirow{1}{*}{\textbf{Vid}}  & \multirow{1}{*}{\textbf{3D (S)}}  &\multirow{1}{*}{\textbf{3D (C)}} &\multirow{1}{*}{\textbf{Text}}\\
			\hline
			GOD \cite{horikawa2017generic}  & \textcolor{red}{\ding{55}} & \textcolor{green}{\ding{51}} & \textcolor{red}{\ding{55}} & \textcolor{green}{\ding{51}} & \textcolor{red}{\ding{55}} & \textcolor{red}{\ding{55}} & \textcolor{red}{\ding{55}}  & \textcolor{red}{\ding{55}} \\
			BOLD5000 \cite{chang2019bold5000}	& \textcolor{red}{\ding{55}} & \textcolor{green}{\ding{51}} & \textcolor{red}{\ding{55}} & \textcolor{green}{\ding{51}} & \textcolor{red}{\ding{55}} & \textcolor{red}{\ding{55}} & \textcolor{red}{\ding{55}}  & \textcolor{red}{\ding{55}} \\
			\selectfont NSD \cite{allen2022massive}	& \textcolor{green}{\ding{51}}  & \textcolor{green}{\ding{51}} & \textcolor{red}{\ding{55}} & \textcolor{green}{\ding{51}} & \textcolor{red}{\ding{55}} & \textcolor{red}{\ding{55}} & \textcolor{red}{\ding{55}}  & \textcolor{green}{\ding{51}} \\
			Video-fMRI \cite{wen2018neural}   & \textcolor{red}{\ding{55}} & \textcolor{green}{\ding{51}} & \textcolor{red}{\ding{55}} & \textcolor{green}{\ding{51}} & \textcolor{green}{\ding{51}} & \textcolor{red}{\ding{55}} & \textcolor{red}{\ding{55}}  & \textcolor{red}{\ding{55}} \\
			Mind-3D \cite{gao2023mind}     & \textcolor{red}{\ding{55}} & \textcolor{green}{\ding{51}} & \textcolor{green}{\ding{51}} & \textcolor{green}{\ding{51}} & \textcolor{green}{\ding{51}} & \textcolor{green}{\ding{51}} & \textcolor{red}{\ding{55}}  & \textcolor{green}{\ding{51}} \\
			\hline
			ImgNet-EEG \cite{kavasidis2017brain2image}   & \textcolor{red}{\ding{55}} & \textcolor{green}{\ding{51}} & \textcolor{red}{\ding{55}} & \textcolor{green}{\ding{51}} & \textcolor{red}{\ding{55}} & \textcolor{red}{\ding{55}} & \textcolor{red}{\ding{55}}  & \textcolor{red}{\ding{55}} \\
			\selectfont Things-EEG \cite{gifford2022large}   & \textcolor{green}{\ding{51}} & \textcolor{green}{\ding{51}} & \textcolor{red}{\ding{55}} & \textcolor{green}{\ding{51}} & \textcolor{red}{\ding{55}} & \textcolor{red}{\ding{55}} & \textcolor{red}{\ding{55}}  & \textcolor{red}{\ding{55}} \\
			\textbf{EEG-3D}   & \textcolor{green}{\ding{51}} & \textcolor{green}{\ding{51}} & \textcolor{green}{\ding{51}} & \textcolor{green}{\ding{51}} &  \textcolor{green}{\ding{51}} &  \textcolor{green}{\ding{51}} &  \textcolor{green}{\ding{51}}  & \textcolor{green}{\ding{51}} \\
			\hline
		\end{tabular}
	}
	\caption{Comparison between \textbf{EEG-3D} and other datasets, categorizing brain activity into resting-state (Re), responses to static stimuli (St) and dynamic stimuli (Dy). The analysis data includes images (Img), videos (Vid), text captions (Text), 3D shape (3D (S)) and color attributes (3D (C)).}
	\label{dataset_comparison}
	\vspace{-0.5cm}
\end{table}

\subsection{Dataset Attributes}

Tab. \ref{dataset_comparison} presents a comparison between \textbf{EEG-3D} and other commonly used datasets \cite{horikawa2017generic,chang2019bold5000,wen2018neural,gao2023mind,kavasidis2017brain2image,gifford2022large,grootswagers2022human,allen2022massive}. Our dataset addresses the gap in the field of extracting 3D information from EEG signals. The EED-3D dataset distinguishes itself from existing datasets by following attributes: 
\begin{itemize}
	\item \textbf{Comprehensive EEG signal recordings.} Our dataset includes resting EEG data, EEG responses to static image stimuli and dynamic video stimuli. These signals enable more comprehensive investigations into neural activity, particularly in understanding the brain's response mechanisms to 3D visual stimuli, as well as comparative analyses of how the visual processing system engages with different types of visual input.
	\item \textbf{Multimodal analysis data and labels.} EEG-3D dataset includes static images, high-resolution videos, text captions and 
	3D shape with color attributes
	aligned with EEG. Each 3D object is annotated with category labels and main color style labels. This comprehensive dataset, with multimodal analysis data and labels, supports a broad range of EEG signal decoding and analysis tasks.
\end{itemize}

These attributes provide a strong basis for exploring brains response mechanisms to dynamic and static stimuli, 
positioning the dataset as a valuable resource for advancing research in neuroscience and computer vision.

	\vspace{-0.1cm}
\section{Method}

\begin{figure*}[t]
	\centering
	\includegraphics[width=0.8\textwidth]{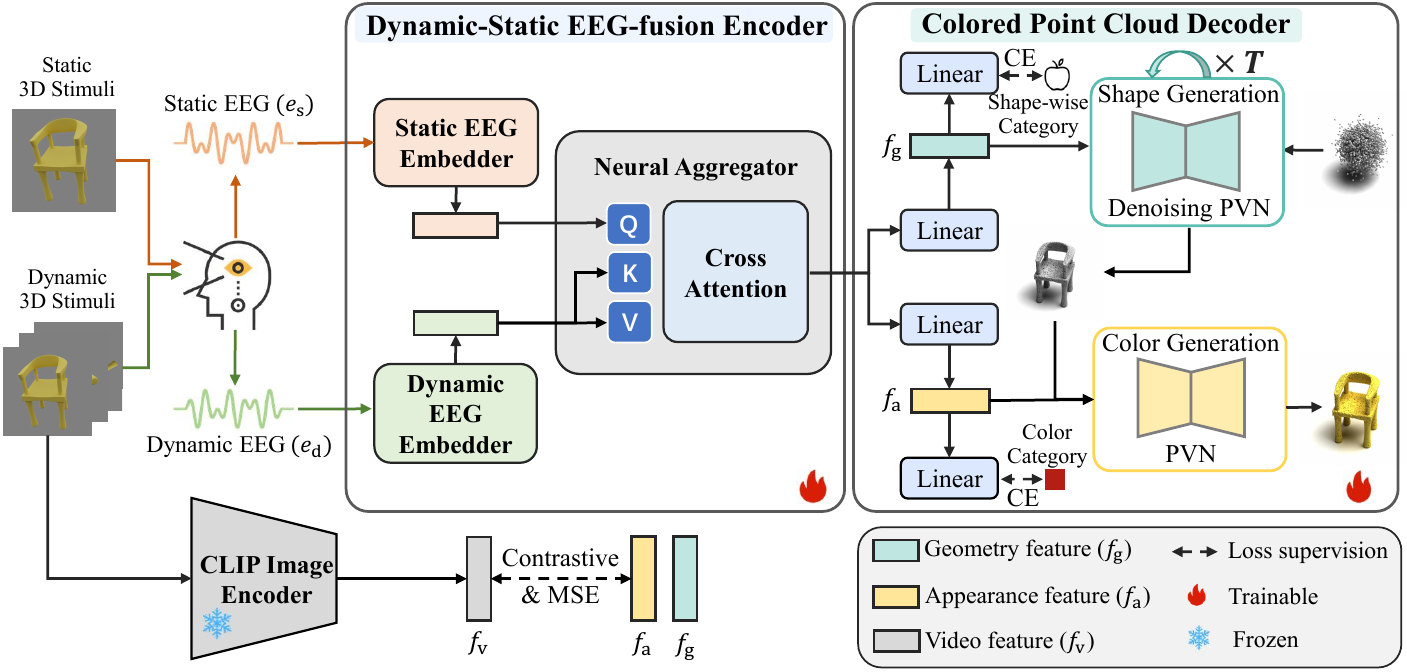} 
	\caption{The proposed Neuro-3D for 3D reconstruction from EEG. The input static and dynamic signals ($e_\mathrm{s}$ and $e_\mathrm{d}$) are aggregated via the dynamic-static EEG-fusion encoder. Subsequently, the fused EEG features are  decoupled into geometry and appearance features ($f_\mathrm{g}$ and $f_\mathrm{a}$). After aligning with clip image embeddings, $f_\mathrm{g}$ and $f_\mathrm{a}$ serve as guidance for the generation of geometric shapes and overall colors.} 
	\label{framework}
	\vspace{-0.3cm}
\end{figure*}

\subsection{Overview}

As depicted in Fig. \ref{framework}, we delineate our framework into two principal components:
1) \textbf{Dynamic-Static EEG-fusion Encoder}: Given the static and dynamic EEG signals ($e_\mathrm{s}$ and $e_\mathrm{d}$) from EEG-3D, the encoder is responsible to extract discriminative neural features by adaptively aggregating dynamic and static EEG features, leveraging their complementary characteristics.
2) \textbf{Colored Point Cloud Decoder}: To reconstruct 3D objects, a two-stage decoder module is proposed to generate 3D shape and color sequentially, conditioned on decoupled geometry and appearance EEG features ($f_\mathrm{g}$ and $f_\mathrm{a}$), respectively. 

\subsection{Dynamic-Static EEG-fusion Encoder}

Given EEG recordings under the static and dynamic 3D visual stimuli, how to extract robust and discriminative neural representations becomes a critical issue.
EEG signals have inherent high noise levels and prolonged exposure to rapidly changing video stimuli introduces further interference factors.
To address this challenge, we propose to adaptively fuse dynamic and static EEG signals for learning comprehensive and robust neural representation.

{\textbf{EEG Embedder.}}
Given preprocessed EEG signals $e_\mathrm{s}\in \mathbb{R}^{C\times T_\mathrm{s}}$ and $e_\mathrm{d}\in \mathbb{R}^{C\times T_\mathrm{d}}$ under static image stimuli $v_0$ (the initial frame of the video) and dynamic video stimuli $\{v_i\}$ with rotational 3D object.
We design two EEG embedders, $E_\mathrm{s}$ and $E_\mathrm{d}$, to extract static and dynamic EEG features from $e_\mathrm{s}$ and $e_\mathrm{d}$, respectively:
\begin{equation}
	z_\mathrm{s}=E_\mathrm{s}(e_\mathrm{s}), z_\mathrm{d}=E_\mathrm{d}(e_\mathrm{d}).
\end{equation}
Specifically,
the embedders consist of multiple temporal self-attention layers 
that apply self-attention \cite{vaswani2017attention} mechanism along the EEG temporal dimension.
They capture and integrate temporal dynamics of brain responses over the duration of the stimulus.
Subsequently, an MLP projection layer is applied to generate output EEG embeddings. 

\textbf{Neural Aggregator.}
The static image stimulus, with a duration of 0.5 seconds, helps the subject capture relatively stable single-view information about the 3D object. 
In contrast, dynamic video stimulation renders a holistic 3D representation with rotating views of the object, but its long duration may introduce additional noise.
To leverage the complementary characteristics, 
we introduce an attention-based neural aggregator to integrate static and dynamic EEG embeddings in an adaptive way. Specifically, query features are derived from static EEG features $z_{\mathrm{s}}$, while key and value features are obtained from dynamic EEG features $z_{\mathrm{d}}$:
\begin{equation}
	Q=W^Qz_\mathrm{s}, K=W^Kz_\mathrm{d}, V=W^Vz_\mathrm{d}.
\end{equation}
The attention-based aggregation can be defined as follows:  
\begin{equation}
	z_{\mathrm{sd}}=\mathrm{Softmax}(\frac{QK^T}{\sqrt{d}})\cdot V,
\end{equation}
where $z_{\mathrm{sd}}$ is the aggregated EEG feature.
The attentive aggregation approach leverages the stability provided by the static image responses and the temporal dependencies inherent in video data, enabling robust and comprehensive neural representation learning against high signal noises.

\subsection{Colored Point Cloud Decoder}

To recover 3D experience from neural representations, we propose a colored point cloud decoder that first generates the shape and then assigns colors to the generated point cloud, conditioned on the decoupled EEG representations.

{\textbf{Decoupled Learning of EEG Features.}}
Directly using the same EEG feature for the two generation stages may result in information interference and redundance.
Therefore, to enable targeted conditioning 
of shape and color generation,   
we learn distinct geometry
and appearance components from EEG embeddings in a decoupled way.
Given the EEG feature $z_\mathrm{sd}$ extracted from EEG-fusion encoder, we decouple it into distinct embeddings for geometry and appearance features ($f_\mathrm{g}$ and $f_\mathrm{a}$) through individual MLP projection layers. 
To learn discriminative and semantically meaningful EEG features, 
we align them with video features $f_\mathrm{v}$ encoded by pre-trained CLIP vision encoder $E_\mathrm{v}$ through a contrastive loss and a MSE loss:
\begin{equation}
	L_{\mathrm{align}}(f, f_\mathrm{v})=\alpha \mathrm{CLIP}(f, f_\mathrm{v})+(1-\alpha)\mathrm{MSE}(f, f_\mathrm{v}),
	\label{alpha}
\end{equation}
\begin{equation}    f_\mathrm{v}={\sum_{i=1}^nE_\mathrm{v}(v_i)}/{n},
\end{equation}
where $f$ represents $f_\mathrm{g}$ or $f_\mathrm{a}$, and $\{v_i\}_{i=1}^n$ denotes downsampled video sequence. 
To enhance the learning of geometry and appearance features, a categorical loss $L_{\mathrm{c}}$ is proposed to ensure the decoupled geometry and appearance features can be correctly classified as ground-truth color and shape categories:
\begin{equation}
	L_{\mathrm{c}}=\mathrm{CE}(\hat{y}_\mathrm{g}, y_\mathrm{g})+\mathrm{CE}(\hat{y}_\mathrm{a}, y_\mathrm{a}),
\end{equation}
where $\hat{y}_\mathrm{g}$ and $\hat{y}_\mathrm{a}$ are the shape and color predictions produced by linear classifiers, $y_\mathrm{g}$ and $ y_\mathrm{a}$ denotes ground-truth labels, and $\mathrm{CE}$ denotes cross-entropy loss.
The final loss $L$ integrates alignment loss and categorical loss:
\begin{equation}
	L=L_{\mathrm{align}}(f_\mathrm{g},f_\mathrm{v})+L_{\mathrm{align}}(f_\mathrm{a},f_\mathrm{v})+\gamma L_{\mathrm{c}}.
	\label{gamma}
\end{equation}
Subsequently, $f_\mathrm{g}$ and $f_\mathrm{a}$ are respectively sent into shape generation and color generation streams for precise brain visual interpretation and reconstruction.

\textbf{Shape Generation.}
The point cloud $X_0\in \mathbb{R}^{N\times 3}$ associated with the stimulus signal is incrementally added noise until it converges to an isotropic Gaussian distribution. The noise addition follows a Markov process, characterized by Gaussian transitions with variance scheduled by hyperparameters $\{\beta_i\}_{t=0}^T$, defined as:
\begin{equation}
	q\left(X_t \mid X_{t-1}\right)=\mathcal{N}\left(\sqrt{1-\beta_t} X_{t-1}, \beta_t \mathbf{I}\right).
\end{equation}
The cumulative noise introduction aligns with the Markov chain assumption, enabling derivation of:
\begin{equation}
	q\left(X_{0: T}\right)=q\left(X_0\right) \prod_{t=1}^T q\left(X_t \mid X_{t-1}\right).
\end{equation}
Our objective is to generate the 3D point cloud based on the geometry EEG features $f_\mathrm{g}$. This is achieved through a reverse diffusion process, which reconstructs corrupted data by modeling the posterior distribution $p_\theta(X_{t-1}|X_t)$ at each diffusion step. The transition from the Gaussian state $X_T$ back to the initial point cloud $X_0$ can be represented as:
\begin{equation}
	p_\theta\left(X_{t-1} \mid X_t, f_\mathrm{g}\right)=\mathcal{N}\left(\mu_\theta\left(X_t, t, f_\mathrm{g}\right), \sigma_t^2 \mathbf{I}\right),
\end{equation}
\begin{equation}
	p_\theta\left(X_{0: T}\right)=p\left(X_T\right) \prod_{t=1}^T p_\theta\left(X_{t-1} \mid X_t, f_\mathrm{g}\right),
	\label{denoised}
\end{equation}
where the parameterized network $\mu_\theta$ is a learnable model to iteratively predict the reverse diffusion steps, ensuring that the generated reverse process closely approximates the forward process. To optimize network training, the diffusion model employs the principle of variational inference, maximizing the variational lower bound of the negative log-likelihood, ultimately yielding a loss function expressed as:
\begin{equation}
	\mathcal{L}_t=\mathbb{E}_{X_0\sim q(X_0)}\mathbb{E}_{\epsilon_t\sim \mathcal{N}(0, \mathbf{I})}\left\|\epsilon_t-\mu_\theta\left(X_t, t, f_\mathrm{g}\right)\right\|^2.
\end{equation}

\textbf{Color Generation.}
Previous researches in point cloud generation suggest that jointly generating geometry and color information often leads to performance degradation and model complexity \cite{melas2023pc2,wu2023sketch}.
Therefore, following the work in \cite{melas2023pc2}, we learn a separate single-step coloring model $h_{\phi}$ to reconstruct object color in addition to object shape. 
Specifically, we use the generated point cloud $\hat{X}_0$ with appearance EEG features $f_\mathrm{a}$ as the condition, and send them to coloring model $h_{\phi}$ to estimate the color of the point cloud.
Due to the limited information provided by EEG signals, predicting distinct colors for each point in a 3D structure presents a significant challenge. As an initial step in addressing this issue,
we simplify the task by aggregating color information from the ground-truth point cloud. Through a majority-voting mechanism, we select dominant colors to represent the entire object, thereby reducing the complexity of the color prediction process.

	\section{Experiments}

\subsection{Experimental Setup}

\textbf{Implementation Details.} We utilize the AdamW optimizer \cite{kingma2014adam} with $\beta=(0.95,0.999)$ and an initial learning rate of $1\times 10^{-3}$. The loss coefficients $\alpha$ and $\gamma$ in Eq.~\eqref{alpha} and Eq.~\eqref{gamma} are set to 0.01 and 0.1, respectively. The dimension of the extracted features ($f_\mathrm{g}$ and $f_\mathrm{a}$) is $1024$. The point cloud consists of $N=8192$ points, and each video sequence is downsampled to $n=4$ frames for feature extraction to facilitate alignment with EEG features. Our method is implemented in PyTorch on a single A100 GPU.  In colored point cloud decoder, Point-Voxel Network (PVN)~\cite{liu2019point} is used as the denoising function of shape diffusion model and single-step color prediction model.

\textbf{Evaluation Benchmarks.} To thoroughly evaluate the 3D decoding performance on EEG-3D, we construct two evaluation benchmarks: 3D visual classification benchmark for evaluating the EEG encoder, and 3D object reconstruction benchmark for assessing the 3D reconstruction pipeline.
(1) \textbf{3D visual classification benchmark.} To assess the high-level visual semantic decoding performance on EEG signals,
we evaluate on two classification tasks: object classification (72 categories) and color type classification (6 categories).
We select top-K accuracies as the evaluation metric for these tasks. 
(2) \textbf{3D object reconstruction benchmark.} 
Following 2D visual decoding methods~\cite{li2024visual,chen2024cinematic,chen2023seeing}, we adopt N-way top-K accuracy to assess the semantic fidelity of generated 3D objects.
Specifically, we train an additional classifier to predict objects categories of point clouds, with training data derived from the Objaverse dataset \cite{deitke2023objaverse, xu2023pointllm}.
The evaluation metrics include 2-way top-1 and 10-way top-3 accuracies, calculated from the average across five generation results as well as the best-performing result in each case. 
Furthermore, we calculate 3D point cloud-based geometric metrics (Chamfer Distance ($\times 10^{-2}$) and F1 score ($\%$)) to assess depth-related information.

\begin{figure*}[t]
	\centering
	\includegraphics[width=0.80\textwidth]{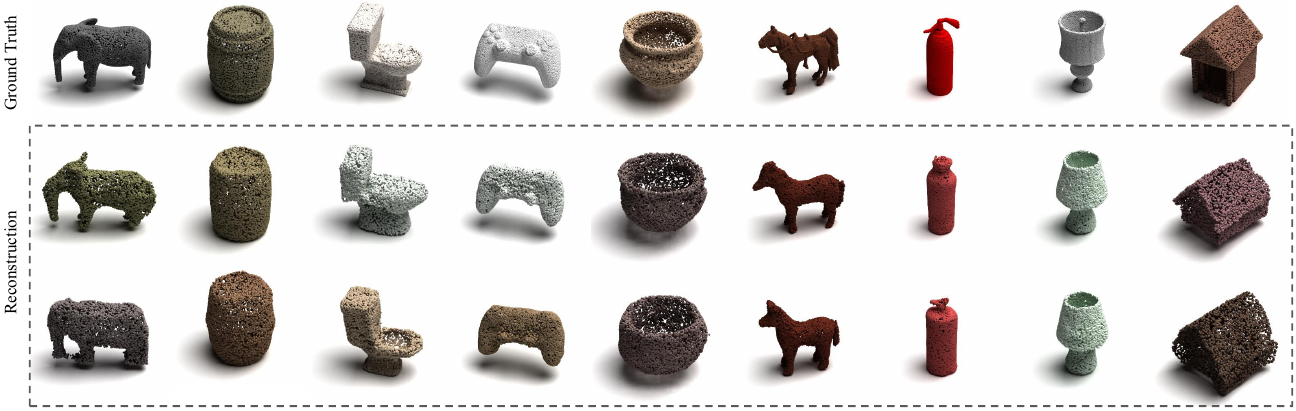} 
	\caption{The qualitative results reconstructed by Neuro-3D with two different samplings trials, and the corresponding ground truth.}
	\label{fig-results}
	\vspace{-0.2cm}
\end{figure*}

\subsection{Classification Task}

We assess the performance of the proposed dynamic-static EEG-fusion encoder on the classification tasks. 

\vspace{-0.2cm}

\subsubsection{Comparison with Related Methods}

We re-implement several state-of-the-art EEG encoders~\cite{schirrmeister2017deep,lawhern2018eegnet,song2022eeg,song2023decoding} for comparative analysis by training separate object and color classifiers. 
Tab. \ref{tab-classify} presents the overall accuracy of various EEG classifiers. All methods exceed chance-level performance by a significant margin, suggesting that the collected EEG signals successfully capture the visual perception processed in the brain. Notably, our proposed EEG-fusion encoder outperforms all baseline methods across all metrics, demonstrating its superior ability in extracting semantically meaningful and discriminative neural representations related with high-level visual perception.

\begin{table}[t]
	\centering
	\renewcommand\arraystretch{1.0}
	\footnotesize
	\begin{tabular}{ c c  c  c  c}
		\Xhline{2\arrayrulewidth}
		
		\multirow{2}*[-3pt]{\textbf{Method}} 
		\rule{0pt}{2pt} & \multicolumn{2}{c}{\multirow{1}{*}[-1pt]{\textbf{\textbf{Object Type}}}}
		& \multicolumn{2}{c}{\multirow{1}{*}[-1pt]{\textbf{\textbf{Color Type}}}} \\ [-1.5pt]
		\cmidrule(l{5pt}r{5pt}){2-3}\cmidrule(l{5pt}r{5pt}){4-5}
		& \multirow{1}{*}[0.8pt]{\textbf{top-1}} & \multirow{1}{*}[0.8pt]{\textbf{top-5}}  & \multirow{1}{*}[0.8pt]{\textbf{top-1}} 
		& \multirow{1}{*}[0.8pt]{\textbf{top-2}}\\
		\hline
		\multirow{1}{*}{Chance level} 
		& 1.39  & 6.94 & 16.67 & 33.33 \\
		\hline
		\multirow{1}{*}{DeepNet (2017) \cite{schirrmeister2017deep}} 
		& 3.70  & 9.90 & 20.95 & 49.71 \\
		\multirow{1}{*}{EEGNet (2018) \cite{lawhern2018eegnet}} 
		& 3.82  & 9.72 & 18.35 & 46.47 \\
		\multirow{1}{*}{Conformer (2023) \cite{song2022eeg}} 
		& 4.05  & 10.30 & 18.27 & 35.81 \\
		\multirow{1}{*}{TSConv (2024) \cite{song2023decoding}} 
		& 4.05  & 10.13 & 31.13 & 59.49 \\ \hline
		\multirow{1}{*}{\textbf{Neuro-3D}} 
		& \textbf{5.91} & \textbf{16.30} & \textbf{39.93} & \textbf{61.40}\\
		\Xhline{2\arrayrulewidth}
	\end{tabular}
	\caption{Comparison results on two classification tasks.}
	\label{tab-classify}
	\vspace{-0.3cm}
\end{table}

\begin{table}[b]
	\centering
	\renewcommand\arraystretch{1.0}
	\footnotesize
	\begin{tabular}{ c c c c c  c  c }
		\Xhline{2\arrayrulewidth}
		
		\multirow{2}*[-3pt]{\textbf{St.}} & \multirow{2}*[-3pt]{\textbf{Dy.}} & \multirow{2}*[-3pt]{\textbf{Agg.}} \rule{0pt}{2pt} & \multicolumn{2}{c}{\multirow{1}{*}[-1pt]{\textbf{\textbf{Object Type}}}}
		& \multicolumn{2}{c}{\multirow{1}{*}[-1pt]{\textbf{\textbf{Color Type}}}} \\ [-1.5pt]
		\cmidrule(l{5pt}r{5pt}){4-5}\cmidrule(l{5pt}r{5pt}){6-7}
		& & & \multirow{1}{*}[0.8pt]{\textbf{top-1}} & \multirow{1}{*}[0.8pt]{\textbf{top-5}}  & \multirow{1}{*}[0.8pt]{\textbf{top-1}} 
		& \multirow{1}{*}[0.8pt]{\textbf{top-2}}\\
		\hline
		\textcolor{green}{\ding{51}} & \textcolor{red}{\ding{55}} & \textcolor{red}{\ding{55}}  & 5.10   & 15.62 & 37.50 & 57.64 \\
		\textcolor{red}{\ding{55}} & \textcolor{green}{\ding{51}} & \textcolor{red}{\ding{55}}  & 4.75 & 13.89 & 35.65 & 55.61 \\ 
		\textcolor{green}{\ding{51}} & \textcolor{green}{\ding{51}} & \textcolor{red}{\ding{55}}  & 5.44& 15.86 & 39.12 & 58.85 \\
		\textcolor{green}{\ding{51}} & \textcolor{green}{\ding{51}} & \textcolor{green}{\ding{51}}  & \textbf{5.91} & \textbf{16.30} & \textbf{39.93} & \textbf{61.40}\\
		\Xhline{2\arrayrulewidth}
	\end{tabular}
	\caption{Ablation experiment results on classification tasks. The table presents the results obtained using static signals (St.), dynamic signals (Dy.), as well as the outcomes from the concatenation of the two signal types without feature aggregation (Agg.).}
	\label{tab-ablation}
\end{table}

\subsubsection{Ablation Study}

We conduct an ablation study to assess the impact of using different EEG signals and modules, as shown in Tab. \ref{tab-ablation}. Compared to using only dynamic features, the performance improves when static features are incorporated. This enhancement may be attributed to the longer duration of the video stimulus, during which factors such as blinking and distraction introduce noise into the dynamic signal, thereby reducing its effectiveness. When the static and dynamic features are simply concatenated, the performance improves compared to using either signal alone, suggesting complementary information between the two signals. Further performance gains are achieved through our attention-based neural aggregator, which adaptively integrates the dynamic and static features. This demonstrates that our method can leverage the information from both EEG features while mitigating the challenges posed by the low signal-to-noise ratio inherent in EEG, thereby enhancing model robustness. 

\begin{table}[]
	\centering
	\renewcommand\arraystretch{1.0}
	\footnotesize
	\begin{tabular}{ c c c  c  c  c  c}
		\Xhline{2\arrayrulewidth}
		
		\multirow{2}*[-3pt]{\textbf{Method}} \rule{0pt}{2pt} & \multicolumn{2}{c}{\multirow{1}{*}[-1pt]{\textbf{\textbf{Average}}}}
		& \multicolumn{4}{c}{\multirow{1}{*}[-1pt]{\textbf{\textbf{Top-1 of 5 samples}}}}  \\ [-1.5pt]
		\cmidrule(l{5pt}r{5pt}){2-3}\cmidrule(l{5pt}r{5pt}){4-7}
		& \multirow{1}{*}[0.8pt]{\textbf{(2, 1)}} & \multirow{1}{*}[0.8pt]{\textbf{(10, 3)}}  & \multirow{1}{*}[0.8pt]{\textbf{(2, 1)}} 
		& \multirow{1}{*}[0.8pt]{\textbf{(10, 3)}} & \multirow{1}{*}[0.8pt]{\textbf{CD}} & \multirow{1}{*}[0.8pt]{\textbf{F1}}\\
		\hline
		\textbf{Static} & 51.64 & 32.39 & 68.75  & 55.14 & 6.75 & 67.61 \\
		\textbf{Dynamic} & 50.86 & 31.50 & 71.25 & 54.30 & 6.40 & 69.42 \\
		\textbf{Concat} & 53.22 & 34.11 & 69.72 & 56.53 & 5.84 & 73.47 \\
		\textbf{w/o De.} & 53.94  & 34.42 & 65.00 & 48.54 & 5.81 & 73.36 \\
		\textbf{Full} & \textbf{55.81}  & \textbf{35.89} & \textbf{72.08} & \textbf{57.64} & \textbf{5.35} & \textbf{77.01} \\
		\Xhline{2\arrayrulewidth}
	\end{tabular}
	\caption{Quantitative results of 3D reconstruction, where (N, t) indicates (N-way top-K) result of reconstructed samples.}
	\label{tab-recon}
	\vspace{-0.2cm}
\end{table}

\subsection{3D Reconstruction Task}

\subsubsection{Quantitative Results}

Quantitative results of various baseline models and our proposed Neuro-3D model are presented in Tab.~\ref{tab-recon}. The generation performance is notably reduced when employing static or dynamic EEG features in isolation.  
Moreover, dynamic information leads to improved performance in geometric metrics compared to static information alone, suggesting that dynamic stimuli provide richer depth and geometric information in EEG signals.
Static EEG features offer stability yet lack sufficient 3D details, whereas dynamic video features provide a more comprehensive 3D representation but suffer from a lower signal-to-noise ratio. 
Integrating static and dynamic features leads to more comprehensive and stable neural representation, thereby enhancing the generation performance.
Furthermore, compared to direct feature concatenation, our proposed neural aggregator effectively merges static and dynamic information, reducing noise interference and further improving reconstruction performance. The decoupling of shape and color features minimizes cross-feature interference, yielding significant advancements in 3D generation quality. Additionally, comparisons between Tab.~\ref{tab-recon} and Tab.~\ref{tab-ablation} reveal a positive correlation between generation quality and classification accuracy, confirming that 
enriching features with high-level semantics enhances visual reconstruction performance.

\subsubsection{Reconstructed Examples}

Fig.~\ref{fig-results} presents the generated results produced by Neuro-3D and the corresponding ground truth objects. The results demonstrate that Neuro-3D not only successfully reconstructs simpler objects such as kegs and potteries but also performs well with more complex structures (such as elephants and horses), underscoring the model’s robust shape perception capabilities. In terms of color generation, while the low spatial resolution of EEG signals poses challenges for detailed texture synthesis, our method effectively captures color styles that closely resemble actual objects. 

\begin{figure}[t]
	\centering
	\begin{subfigure}{0.85\linewidth}
		\centering
		\includegraphics[width=\linewidth]{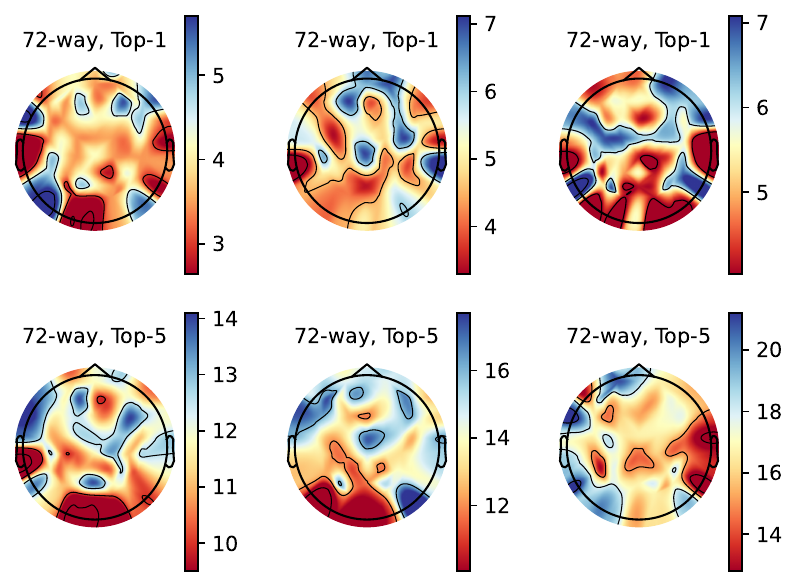} 
		\caption{EEG Electrode Channel Analysis}
		\label{fig_areas1}
	\end{subfigure}

	\begin{subfigure}{0.85\linewidth}
		\centering
		\includegraphics[width=\linewidth]{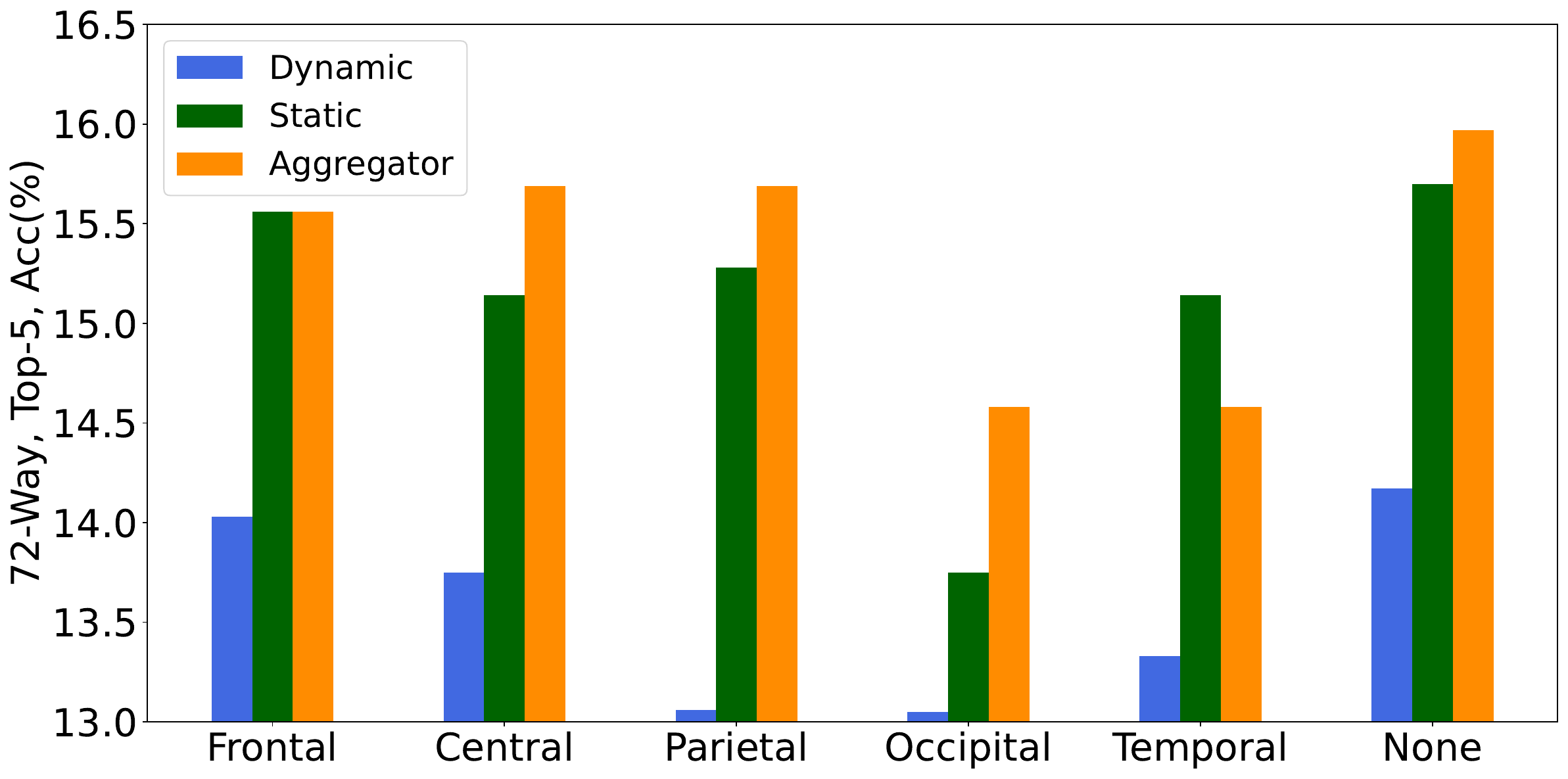}
		\caption{Brain Region Analysis}
		\label{fig_areas2}
	\end{subfigure}
	\caption{Analysis of brain areas. (a) displays the top-1 and top-5 accuracies of 3 subjects on object classification task by removing individual EEG electrode channels. (b) illustrates the top-5 classification results after selectively removing electrodes from different brain regions, under varying signal conditions.}
	\label{fig_areas}
	\vspace{-0.2cm}
\end{figure}

\subsection{Analysis of Brain Regions}
\label{analysis-brain}

To examine the contribution of different brain regions to 3D visual perception, we generated saliency maps for 3 subjects by sequentially removing each of the 64 electrode channels, as illustrated in Fig. \ref{fig_areas}~(a). Notably, the removal of occipital electrodes presents the most significant effect on performance, as this region is strongly linked to the brain's visual processing pathways. This finding aligns with the previous neuroscience discoveries regarding the brain's visual processing mechanisms~\cite{grill2001lateral,song2023decoding,li2024visual}. Moreover, previous studies have identified the inferior temporal cortex in the temporal lobe as crucial for high-level semantic processing and object recognition~\cite{dicarlo2007untangling,bao2020map}. Consistent with this, the results shown in Fig.~\ref{fig_areas}~(a) suggest a potential correlation between visual decoding performance and this brain region. A comparative analysis of classification results across different subjects reveals substantial variability in EEG signals between individuals. 

We further assess the visual decoding performance by sequentially removing electrodes from five distinct brain regions. As shown in Fig.~\ref{fig_areas}~(b), removal of electrodes from the occipital or temporal regions led to a marked decrease in performance, which is consistent with our expectations. Additionally, removing electrodes from the temporal or parietal regions results in a more pronounced performance decline for dynamic stimuli, compared to static stimuli. This effect is likely attributed to the involvement of the dorsal visual pathway, which is responsible for motion perception and runs from middle temporal visual area, medial superior temporal area and ventral intraparietal cortex in the parietal lobe \cite{salzman1992microstimulation,gu2012causal,chen2011representation}.

	\section{Discussion and Conclusion}
\textbf{Limitations and Future Work}. A limitation of our study is the simplification of texture generation to the main color style prediction due to the complexities of detailed texture synthesis. Extending this work to generate complete 3D textures is a key focus for future research. Moreover, given the substantial individual variations in EEG, future work should extend to enhance cross-subject generalization.\\ 
\textbf{Conclusion}. We explore a new task of reconstructing colored 3D objects from EEG signals, which is challenging but holds considerable importance for understanding the brain's mechanisms for real-time 3D perception. To facilitate this task, we develop the EEG-3D dataset, which integrates multimodal data and extensive EEG recordings. This dataset addresses the scarcity of EEG-3D object pairings, providing a valuable resource for future research in this domain. Furthermore, we propose a new framework, Neuro-3D, for extracting EEG-based visual features and reconstructing 3D objects. Neuro-3D leverages a diffusion-based 3D decoder for shape and color generation, conditioned on adaptively fused EEG features captured under static and dynamic 3D stimuli.
Extensive experiments demonstrate the feasibility of decoding 3D information from EEG signals and confirms the alignment between EEG visual decoding and biological visual perception mechanisms.
	\section*{Acknowledgments}
	This work is supported by Shanghai Artificial Intelligence Laboratory. 
	\newpage
	{
		\small
		\bibliographystyle{ieeenat_fullname}
		\bibliography{EEGTo3D}

\begin{thebibliography}{83}
\providecommand{\natexlab}[1]{#1}
\providecommand{\url}[1]{\texttt{#1}}
\expandafter\ifx\csname urlstyle\endcsname\relax
  \providecommand{\doi}[1]{doi: #1}\else
  \providecommand{\doi}{doi: \begingroup \urlstyle{rm}\Url}\fi

\bibitem[Allen et~al.(2022)Allen, St-Yves, Wu, Breedlove, Prince, Dowdle, Nau,
  Caron, Pestilli, Charest, et~al.]{allen2022massive}
Emily~J Allen, Ghislain St-Yves, Yihan Wu, Jesse~L Breedlove, Jacob~S Prince,
  Logan~T Dowdle, Matthias Nau, Brad Caron, Franco Pestilli, Ian Charest,
  et~al.
\newblock A massive 7t fmri dataset to bridge cognitive neuroscience and
  artificial intelligence.
\newblock \emph{Nature Neuroscience}, 25\penalty0 (1):\penalty0 116--126, 2022.

\bibitem[Bai et~al.(2023)Bai, Wang, Cao, Ge, Yuan, and
  Shan]{bai2023dreamdiffusion}
Yunpeng Bai, Xintao Wang, Yan-pei Cao, Yixiao Ge, Chun Yuan, and Ying Shan.
\newblock Dreamdiffusion: Generating high-quality images from brain eeg
  signals.
\newblock \emph{arXiv preprint arXiv:2306.16934}, 2023.

\bibitem[Bao et~al.(2020)Bao, She, McGill, and Tsao]{bao2020map}
Pinglei Bao, Liang She, Mason McGill, and Doris~Y Tsao.
\newblock A map of object space in primate inferotemporal cortex.
\newblock \emph{Nature}, 583\penalty0 (7814):\penalty0 103--108, 2020.

\bibitem[Beliy et~al.(2019)Beliy, Gaziv, Hoogi, Strappini, Golan, and
  Irani]{beliy2019voxels}
Roman Beliy, Guy Gaziv, Assaf Hoogi, Francesca Strappini, Tal Golan, and Michal
  Irani.
\newblock From voxels to pixels and back: {S}elf-supervision in natural-image
  reconstruction from {fMRI}.
\newblock \emph{Advances in Neural Information Processing Systems}, 32, 2019.

\bibitem[Chang et~al.(2019)Chang, Pyles, Marcus, Gupta, Tarr, and
  Aminoff]{chang2019bold5000}
Nadine Chang, John~A Pyles, Austin Marcus, Abhinav Gupta, Michael~J Tarr, and
  Elissa~M Aminoff.
\newblock {BOLD5000}, a public {fMRI} dataset while viewing 5000 visual images.
\newblock \emph{Scientific Data}, 6\penalty0 (1):\penalty0 49, 2019.

\bibitem[Chen et~al.(2011)Chen, DeAngelis, and
  Angelaki]{chen2011representation}
Aihua Chen, Gregory~C DeAngelis, and Dora~E Angelaki.
\newblock Representation of vestibular and visual cues to self-motion in
  ventral intraparietal cortex.
\newblock \emph{Journal of Neuroscience}, 31\penalty0 (33):\penalty0
  12036--12052, 2011.

\bibitem[Chen et~al.(2024{\natexlab{a}})Chen, Ren, Song, Wang, Wang, Li, and
  Qiu]{chen2024eegformer}
Yuqi Chen, Kan Ren, Kaitao Song, Yansen Wang, Yifan Wang, Dongsheng Li, and
  Lili Qiu.
\newblock Eegformer: Towards transferable and interpretable large-scale eeg
  foundation model.
\newblock \emph{arXiv preprint arXiv:2401.10278}, 2024{\natexlab{a}}.

\bibitem[Chen et~al.(2023)Chen, Qing, Xiang, Yue, and Zhou]{chen2023seeing}
Zijiao Chen, Jiaxin Qing, Tiange Xiang, Wan~Lin Yue, and Juan~Helen Zhou.
\newblock Seeing beyond the brain: {M}asked modeling conditioned diffusion
  model for human vision decoding.
\newblock In \emph{Proceedings of the IEEE/CVF Conference on Computer Vision
  and Pattern Recognition}, 2023.

\bibitem[Chen et~al.(2024{\natexlab{b}})Chen, Qing, and
  Zhou]{chen2024cinematic}
Zijiao Chen, Jiaxin Qing, and Juan~Helen Zhou.
\newblock Cinematic mindscapes: High-quality video reconstruction from brain
  activity.
\newblock \emph{Advances in Neural Information Processing Systems}, 36,
  2024{\natexlab{b}}.

\bibitem[Deitke et~al.(2023)Deitke, Schwenk, Salvador, Weihs, Michel,
  VanderBilt, Schmidt, Ehsani, Kembhavi, and Farhadi]{deitke2023objaverse}
Matt Deitke, Dustin Schwenk, Jordi Salvador, Luca Weihs, Oscar Michel, Eli
  VanderBilt, Ludwig Schmidt, Kiana Ehsani, Aniruddha Kembhavi, and Ali
  Farhadi.
\newblock Objaverse: {A} universe of annotated {3D} objects.
\newblock In \emph{Proceedings of the IEEE/CVF Conference on Computer Vision
  and Pattern Recognition}, pages 13142--13153, 2023.

\bibitem[DiCarlo and Cox(2007)]{dicarlo2007untangling}
James~J DiCarlo and David~D Cox.
\newblock Untangling invariant object recognition.
\newblock \emph{Trends in Cognitive Sciences}, 11\penalty0 (8):\penalty0
  333--341, 2007.

\bibitem[Engel et~al.(1994)Engel, Rumelhart, Wandell, Lee, Glover,
  Chichilnisky, Shadlen, et~al.]{engel1994fmri}
Stephen~A Engel, David~E Rumelhart, Brian~A Wandell, Adrian~T Lee, Gary~H
  Glover, Eduardo-Jose Chichilnisky, Michael~N Shadlen, et~al.
\newblock {fMRI} of human visual cortex.
\newblock \emph{Nature}, 369\penalty0 (6481):\penalty0 525--525, 1994.

\bibitem[Gao et~al.(2024{\natexlab{a}})Gao, Fu, Wang, Qian, Feng, and
  Fu]{fMRI-3D}
Jianxiong Gao, Yuqian Fu, Yun Wang, Xuelin Qian, Jianfeng Feng, and Yanwei Fu.
\newblock fmri-3d: A comprehensive dataset for enhancing fmri-based 3d
  reconstruction.
\newblock \emph{arXiv preprint arXiv:2409.11315}, 2024{\natexlab{a}}.

\bibitem[Gao et~al.(2024{\natexlab{b}})Gao, Fu, Wang, Qian, Feng, and
  Fu]{gao2023mind}
Jianxiong Gao, Yuqian Fu, Yun Wang, Xuelin Qian, Jianfeng Feng, and Yanwei Fu.
\newblock {Mind-3D}: {R}econstruct high-quality {3D} objects in human brain.
\newblock In \emph{European Conference on Computer Vision}, 2024{\natexlab{b}}.

\bibitem[Gibson et~al.(2022)Gibson, Lobaugh, Joordens, and
  McIntosh]{gibson2022eeg}
Erin Gibson, Nancy~J Lobaugh, Steve Joordens, and Anthony~R McIntosh.
\newblock Eeg variability: Task-driven or subject-driven signal of interest?
\newblock \emph{NeuroImage}, 252:\penalty0 119034, 2022.

\bibitem[Gifford et~al.(2022)Gifford, Dwivedi, Roig, and
  Cichy]{gifford2022large}
Alessandro~T Gifford, Kshitij Dwivedi, Gemma Roig, and Radoslaw~M Cichy.
\newblock A large and rich {EEG} dataset for modeling human visual object
  recognition.
\newblock \emph{NeuroImage}, 264:\penalty0 119754, 2022.

\bibitem[Gramfort et~al.(2013)Gramfort, Luessi, Larson, Engemann, Strohmeier,
  Brodbeck, Goj, Jas, Brooks, Parkkonen, et~al.]{gramfort2013meg}
Alexandre Gramfort, Martin Luessi, Eric Larson, Denis~A Engemann, Daniel
  Strohmeier, Christian Brodbeck, Roman Goj, Mainak Jas, Teon Brooks, Lauri
  Parkkonen, et~al.
\newblock {MEG} and {EEG} data analysis with {MNE-Python}.
\newblock \emph{Frontiers in Neuroinformatics}, 7:\penalty0 267, 2013.

\bibitem[Grill-Spector and Malach(2004)]{grill2004human}
Kalanit Grill-Spector and Rafael Malach.
\newblock The human visual cortex.
\newblock \emph{Annual Review of Neuroscience}, 27\penalty0 (1):\penalty0
  649--677, 2004.

\bibitem[Grill-Spector et~al.(2001)Grill-Spector, Kourtzi, and
  Kanwisher]{grill2001lateral}
Kalanit Grill-Spector, Zoe Kourtzi, and Nancy Kanwisher.
\newblock The lateral occipital complex and its role in object recognition.
\newblock \emph{Vision Research}, 41\penalty0 (10-11):\penalty0 1409--1422,
  2001.

\bibitem[Grootswagers et~al.(2022)Grootswagers, Zhou, Robinson, Hebart, and
  Carlson]{grootswagers2022human}
Tijl Grootswagers, Ivy Zhou, Amanda~K Robinson, Martin~N Hebart, and Thomas~A
  Carlson.
\newblock Human {EEG} recordings for 1,854 concepts presented in rapid serial
  visual presentation streams.
\newblock \emph{Scientific Data}, 9\penalty0 (1):\penalty0 3, 2022.

\bibitem[Gu et~al.(2012)Gu, DeAngelis, and Angelaki]{gu2012causal}
Yong Gu, Gregory~C DeAngelis, and Dora~E Angelaki.
\newblock Causal links between dorsal medial superior temporal area neurons and
  multisensory heading perception.
\newblock \emph{Journal of Neuroscience}, 32\penalty0 (7):\penalty0 2299--2313,
  2012.

\bibitem[Guggenmos et~al.(2018)Guggenmos, Sterzer, and
  Cichy]{guggenmos2018multivariate}
Matthias Guggenmos, Philipp Sterzer, and Radoslaw~Martin Cichy.
\newblock Multivariate pattern analysis for meg: A comparison of dissimilarity
  measures.
\newblock \emph{NeuroImage}, 173:\penalty0 434--447, 2018.

\bibitem[Hebb(2005)]{hebb2005organization}
Donald~Olding Hebb.
\newblock \emph{The organization of behavior: A neuropsychological theory}.
\newblock Psychology press, 2005.

\bibitem[Hendee and Wells(1997)]{hendee1997perception}
William~R Hendee and Peter~NT Wells.
\newblock \emph{The perception of visual information}.
\newblock Springer Science \& Business Media, 1997.

\bibitem[Ho et~al.(2020)Ho, Jain, and Abbeel]{ho2020denoising}
Jonathan Ho, Ajay Jain, and Pieter Abbeel.
\newblock Denoising diffusion probabilistic models.
\newblock \emph{Advances in Neural Information Processing Systems},
  33:\penalty0 6840--6851, 2020.

\bibitem[Horikawa and Kamitani(2017)]{horikawa2017generic}
Tomoyasu Horikawa and Yukiyasu Kamitani.
\newblock Generic decoding of seen and imagined objects using hierarchical
  visual features.
\newblock \emph{Nature Communications}, 8\penalty0 (1):\penalty0 15037, 2017.

\bibitem[Huang et~al.(2023)Huang, Zhao, Zhang, Hu, Fan, Fu, Chen, Xiao, Wang,
  and Dan]{huang2023discrepancy}
Gan Huang, Zhiheng Zhao, Shaorong Zhang, Zhenxing Hu, Jiaming Fan, Meisong Fu,
  Jiale Chen, Yaqiong Xiao, Jun Wang, and Guo Dan.
\newblock Discrepancy between inter-and intra-subject variability in eeg-based
  motor imagery brain-computer interface: Evidence from multiple perspectives.
\newblock \emph{Frontiers in Neuroscience}, 17:\penalty0 1122661, 2023.

\bibitem[Huang et~al.(2021)Huang, Shao, Wang, and Zhang]{huang2021fmri}
Shuo Huang, Wei Shao, Mei-Ling Wang, and Dao-Qiang Zhang.
\newblock fmri-based decoding of visual information from human brain activity:
  A brief review.
\newblock \emph{Machine Intelligence Research}, 18\penalty0 (2):\penalty0
  170--184, 2021.

\bibitem[Jiang et~al.(2024)Jiang, Zhao, and Lu]{jiang2024large}
Wei-Bang Jiang, Li-Ming Zhao, and Bao-Liang Lu.
\newblock Large brain model for learning generic representations with
  tremendous eeg data in bci.
\newblock In \emph{International Conference on Learning Representations}, 2024.

\bibitem[Kavasidis et~al.(2017)Kavasidis, Palazzo, Spampinato, Giordano, and
  Shah]{kavasidis2017brain2image}
Isaak Kavasidis, Simone Palazzo, Concetto Spampinato, Daniela Giordano, and
  Mubarak Shah.
\newblock Brain2image: {C}onverting brain signals into images.
\newblock In \emph{Proceedings of the 25th ACM International Conference on
  Multimedia}, pages 1809--1817, 2017.

\bibitem[Kim et~al.(2022)Kim, Kwon, and Ye]{kim2022diffusionclip}
Gwanghyun Kim, Taesung Kwon, and Jong~Chul Ye.
\newblock Diffusionclip: {T}ext-guided diffusion models for robust image
  manipulation.
\newblock In \emph{Proceedings of the IEEE/CVF Conference on Computer Vision
  and Pattern Recognition}, pages 2426--2435, 2022.

\bibitem[Kingma and Ba(2014)]{kingma2014adam}
Diederik~P Kingma and Jimmy Ba.
\newblock Adam: A method for stochastic optimization.
\newblock \emph{arXiv preprint arXiv:1412.6980}, 2014.

\bibitem[Kupershmidt et~al.(2022)Kupershmidt, Beliy, Gaziv, and
  Irani]{kupershmidt2022penny}
Ganit Kupershmidt, Roman Beliy, Guy Gaziv, and Michal Irani.
\newblock A penny for your (visual) thoughts: {S}elf-supervised reconstruction
  of natural movies from brain activity.
\newblock \emph{arXiv preprint arXiv:2206.03544}, 2022.

\bibitem[Lahner et~al.(2024)Lahner, Dwivedi, Iamshchinina, Graumann, Lascelles,
  Roig, Gifford, Pan, Jin, Ratan~Murty, et~al.]{lahner2024modeling}
Benjamin Lahner, Kshitij Dwivedi, Polina Iamshchinina, Monika Graumann, Alex
  Lascelles, Gemma Roig, Alessandro~Thomas Gifford, Bowen Pan, SouYoung Jin,
  N~Apurva Ratan~Murty, et~al.
\newblock Modeling short visual events through the bold moments video fmri
  dataset and metadata.
\newblock \emph{Nature Communications}, 15\penalty0 (1):\penalty0 6241, 2024.

\bibitem[Lawhern et~al.(2018)Lawhern, Solon, Waytowich, Gordon, Hung, and
  Lance]{lawhern2018eegnet}
Vernon~J Lawhern, Amelia~J Solon, Nicholas~R Waytowich, Stephen~M Gordon,
  Chou~P Hung, and Brent~J Lance.
\newblock {EEGNet}: a compact convolutional neural network for {EEG}-based
  brain--computer interfaces.
\newblock \emph{Journal of Neural Engineering}, 15\penalty0 (5):\penalty0
  056013, 2018.

\bibitem[Li et~al.(2024)Li, Wei, Li, Zou, and Liu]{li2024visual}
Dongyang Li, Chen Wei, Shiying Li, Jiachen Zou, and Quanying Liu.
\newblock Visual decoding and reconstruction via {EEG} embeddings with guided
  diffusion.
\newblock In \emph{Advances in Neural Information Processing Systems}, 2024.

\bibitem[Li et~al.(2023)Li, Li, Savarese, and Hoi]{li2023blip}
Junnan Li, Dongxu Li, Silvio Savarese, and Steven Hoi.
\newblock Blip-2: Bootstrapping language-image pre-training with frozen image
  encoders and large language models.
\newblock In \emph{International Conference on Machine Learning}, pages
  19730--19742. PMLR, 2023.

\bibitem[Lin et~al.(2022)Lin, Sprague, and Singh]{lin2022mind}
Sikun Lin, Thomas Sprague, and Ambuj~K Singh.
\newblock Mind reader: {R}econstructing complex images from brain activities.
\newblock \emph{Advances in Neural Information Processing Systems},
  35:\penalty0 29624--29636, 2022.

\bibitem[Liu et~al.(2023)Liu, Wu, Van~Hoorick, Tokmakov, Zakharov, and
  Vondrick]{liu2023zero}
Ruoshi Liu, Rundi Wu, Basile Van~Hoorick, Pavel Tokmakov, Sergey Zakharov, and
  Carl Vondrick.
\newblock Zero-1-to-3: {Z}ero-shot one image to {3D} object.
\newblock In \emph{Proceedings of the IEEE/CVF International Conference on
  Computer Vision}, pages 9298--9309, 2023.

\bibitem[Liu et~al.(2019)Liu, Tang, Lin, and Han]{liu2019point}
Zhijian Liu, Haotian Tang, Yujun Lin, and Song Han.
\newblock Point-voxel cnn for efficient 3d deep learning.
\newblock \emph{Advances in Neural Information Processing Systems}, 32, 2019.

\bibitem[Long et~al.(2024)Long, Guo, Lin, Liu, Dou, Liu, Ma, Zhang, Habermann,
  Theobalt, et~al.]{long2024wonder3d}
Xiaoxiao Long, Yuan-Chen Guo, Cheng Lin, Yuan Liu, Zhiyang Dou, Lingjie Liu,
  Yuexin Ma, Song-Hai Zhang, Marc Habermann, Christian Theobalt, et~al.
\newblock {Wonder3D}: Single image to {3D} using cross-domain diffusion.
\newblock In \emph{Proceedings of the IEEE/CVF Conference on Computer Vision
  and Pattern Recognition}, pages 9970--9980, 2024.

\bibitem[Luo et~al.(2024)Luo, Henderson, Wehbe, and Tarr]{luo2024brain}
Andrew Luo, Maggie Henderson, Leila Wehbe, and Michael Tarr.
\newblock Brain diffusion for visual exploration: Cortical discovery using
  large scale generative models.
\newblock \emph{Advances in Neural Information Processing Systems}, 36, 2024.

\bibitem[Luo and Hu(2021)]{luo2021diffusion}
Shitong Luo and Wei Hu.
\newblock Diffusion probabilistic models for {3D} point cloud generation.
\newblock In \emph{Proceedings of the IEEE/CVF Conference on Computer Vision
  and Pattern Recognition}, pages 2837--2845, 2021.

\bibitem[Melas-Kyriazi et~al.(2023)Melas-Kyriazi, Rupprecht, and
  Vedaldi]{melas2023pc2}
Luke Melas-Kyriazi, Christian Rupprecht, and Andrea Vedaldi.
\newblock Pc2: {P}rojection-conditioned point cloud diffusion for single-image
  {3D} reconstruction.
\newblock In \emph{Proceedings of the IEEE/CVF Conference on Computer Vision
  and Pattern Recognition}, pages 12923--12932, 2023.

\bibitem[Metzger et~al.(2023)Metzger, Littlejohn, Silva, Moses, Seaton, Wang,
  Dougherty, Liu, Wu, Berger, et~al.]{metzger2023high}
Sean~L Metzger, Kaylo~T Littlejohn, Alexander~B Silva, David~A Moses,
  Margaret~P Seaton, Ran Wang, Maximilian~E Dougherty, Jessie~R Liu, Peter Wu,
  Michael~A Berger, et~al.
\newblock A high-performance neuroprosthesis for speech decoding and avatar
  control.
\newblock \emph{Nature}, 620\penalty0 (7976):\penalty0 1037--1046, 2023.

\bibitem[Moses et~al.(2021)Moses, Metzger, Liu, Anumanchipalli, Makin, Sun,
  Chartier, Dougherty, Liu, Abrams, et~al.]{moses2021neuroprosthesis}
David~A Moses, Sean~L Metzger, Jessie~R Liu, Gopala~K Anumanchipalli, Joseph~G
  Makin, Pengfei~F Sun, Josh Chartier, Maximilian~E Dougherty, Patricia~M Liu,
  Gary~M Abrams, et~al.
\newblock Neuroprosthesis for decoding speech in a paralyzed person with
  anarthria.
\newblock \emph{New England Journal of Medicine}, 385\penalty0 (3):\penalty0
  217--227, 2021.

\bibitem[Nichol et~al.(2022)Nichol, Jun, Dhariwal, Mishkin, and
  Chen]{nichol2022point}
Alex Nichol, Heewoo Jun, Prafulla Dhariwal, Pamela Mishkin, and Mark Chen.
\newblock Point-e: A system for generating {3D} point clouds from complex
  prompts.
\newblock \emph{arXiv preprint arXiv:2212.08751}, 2022.

\bibitem[Nichol and Dhariwal(2021)]{nichol2021improved}
Alexander~Quinn Nichol and Prafulla Dhariwal.
\newblock Improved denoising diffusion probabilistic models.
\newblock In \emph{International Conference on Machine Learning}, pages
  8162--8171. PMLR, 2021.

\bibitem[Ozcelik et~al.(2022)Ozcelik, Choksi, Mozafari, Reddy, and
  VanRullen]{ozcelik2022reconstruction}
Furkan Ozcelik, Bhavin Choksi, Milad Mozafari, Leila Reddy, and Rufin
  VanRullen.
\newblock Reconstruction of perceived images from {fMRI} patterns and semantic
  brain exploration using instance-conditioned {GANs}.
\newblock In \emph{2022 International Joint Conference on Neural Networks},
  pages 1--8. IEEE, 2022.

\bibitem[Poole et~al.(2022)Poole, Jain, Barron, and
  Mildenhall]{poole2022dreamfusion}
Ben Poole, Ajay Jain, Jonathan~T Barron, and Ben Mildenhall.
\newblock Dreamfusion: {T}ext-to-{3D} using {2D} diffusion.
\newblock \emph{arXiv preprint arXiv:2209.14988}, 2022.

\bibitem[Qi et~al.(2017)Qi, Yi, Su, and Guibas]{qi2017pointnet++}
Charles~Ruizhongtai Qi, Li Yi, Hao Su, and Leonidas~J Guibas.
\newblock Pointnet++: Deep hierarchical feature learning on point sets in a
  metric space.
\newblock \emph{Advances in Neural Information Processing Systems}, 30, 2017.

\bibitem[Radford et~al.(2021)Radford, Kim, Hallacy, Ramesh, Goh, Agarwal,
  Sastry, Askell, Mishkin, Clark, et~al.]{radford2021learning}
Alec Radford, Jong~Wook Kim, Chris Hallacy, Aditya Ramesh, Gabriel Goh,
  Sandhini Agarwal, Girish Sastry, Amanda Askell, Pamela Mishkin, Jack Clark,
  et~al.
\newblock Learning transferable visual models from natural language
  supervision.
\newblock In \emph{International Conference on Machine Learning}, pages
  8748--8763. PMLR, 2021.

\bibitem[Ramesh et~al.(2021)Ramesh, Pavlov, Goh, Gray, Voss, Radford, Chen, and
  Sutskever]{ramesh2021zero}
Aditya Ramesh, Mikhail Pavlov, Gabriel Goh, Scott Gray, Chelsea Voss, Alec
  Radford, Mark Chen, and Ilya Sutskever.
\newblock Zero-shot text-to-image generation.
\newblock In \emph{International Conference on Machine Learning}, pages
  8821--8831. PMLR, 2021.

\bibitem[Ren et~al.(2024)Ren, Kim, Liu, and Liu]{ren2024tiger}
Zhiyuan Ren, Minchul Kim, Feng Liu, and Xiaoming Liu.
\newblock {TIGER}: {T}ime-varying denoising model for {3D} point cloud
  generation with diffusion process.
\newblock In \emph{Proceedings of the IEEE/CVF Conference on Computer Vision
  and Pattern Recognition}, pages 9462--9471, 2024.

\bibitem[Rombach et~al.(2022)Rombach, Blattmann, Lorenz, Esser, and
  Ommer]{rombach2022high}
Robin Rombach, Andreas Blattmann, Dominik Lorenz, Patrick Esser, and Bj{\"o}rn
  Ommer.
\newblock High-resolution image synthesis with latent diffusion models.
\newblock In \emph{Proceedings of the IEEE/CVF Conference on Computer Vision
  and Pattern Recognition}, pages 10684--10695, 2022.

\bibitem[Saha and Baumert(2020)]{saha2020intra}
Simanto Saha and Mathias Baumert.
\newblock Intra-and inter-subject variability in eeg-based sensorimotor brain
  computer interface: a review.
\newblock \emph{Frontiers in Computational Neuroscience}, 13:\penalty0 87,
  2020.

\bibitem[Saharia et~al.(2022)Saharia, Chan, Saxena, Li, Whang, Denton,
  Ghasemipour, Gontijo~Lopes, Karagol~Ayan, Salimans,
  et~al.]{saharia2022photorealistic}
Chitwan Saharia, William Chan, Saurabh Saxena, Lala Li, Jay Whang, Emily~L
  Denton, Kamyar Ghasemipour, Raphael Gontijo~Lopes, Burcu Karagol~Ayan, Tim
  Salimans, et~al.
\newblock Photorealistic text-to-image diffusion models with deep language
  understanding.
\newblock \emph{Advances in Neural Information Processing Systems},
  35:\penalty0 36479--36494, 2022.

\bibitem[Salzman et~al.(1992)Salzman, Murasugi, Britten, and
  Newsome]{salzman1992microstimulation}
C~Daniel Salzman, Chieko~M Murasugi, Kenneth~H Britten, and William~T Newsome.
\newblock Microstimulation in visual area mt: effects on direction
  discrimination performance.
\newblock \emph{Journal of Neuroscience}, 12\penalty0 (6):\penalty0 2331--2355,
  1992.

\bibitem[Sargent et~al.(2024)Sargent, Li, Shah, Herrmann, Yu, Zhang, Chan,
  Lagun, Fei-Fei, Sun, et~al.]{sargent2023zeronvs}
Kyle Sargent, Zizhang Li, Tanmay Shah, Charles Herrmann, Hong-Xing Yu, Yunzhi
  Zhang, Eric~Ryan Chan, Dmitry Lagun, Li Fei-Fei, Deqing Sun, et~al.
\newblock Zeronvs: {Z}ero-shot 360-degree view synthesis from a single real
  image.
\newblock In \emph{Proceedings of the IEEE/CVF Conference on Computer Vision
  and Pattern Recognition}, pages 9420--9429, 2024.

\bibitem[Schirrmeister et~al.(2017)Schirrmeister, Springenberg, Fiederer,
  Glasstetter, Eggensperger, Tangermann, Hutter, Burgard, and
  Ball]{schirrmeister2017deep}
Robin~Tibor Schirrmeister, Jost~Tobias Springenberg, Lukas Dominique~Josef
  Fiederer, Martin Glasstetter, Katharina Eggensperger, Michael Tangermann,
  Frank Hutter, Wolfram Burgard, and Tonio Ball.
\newblock Deep learning with convolutional neural networks for eeg decoding and
  visualization.
\newblock \emph{Human Mrain Mapping}, 38\penalty0 (11):\penalty0 5391--5420,
  2017.

\bibitem[Scotti et~al.(2024{\natexlab{a}})Scotti, Banerjee, Goode, Shabalin,
  Nguyen, Dempster, Verlinde, Yundler, Weisberg, Norman,
  et~al.]{scotti2024reconstructing}
Paul Scotti, Atmadeep Banerjee, Jimmie Goode, Stepan Shabalin, Alex Nguyen,
  Aidan Dempster, Nathalie Verlinde, Elad Yundler, David Weisberg, Kenneth
  Norman, et~al.
\newblock Reconstructing the mind's eye: {fMRI}-to-image with contrastive
  learning and diffusion priors.
\newblock \emph{Advances in Neural Information Processing Systems}, 36,
  2024{\natexlab{a}}.

\bibitem[Scotti et~al.(2024{\natexlab{b}})Scotti, Tripathy, Villanueva,
  Kneeland, Chen, Narang, Santhirasegaran, Xu, Naselaris, Norman,
  et~al.]{scotti2024mindeye2}
Paul~S Scotti, Mihir Tripathy, Cesar Kadir~Torrico Villanueva, Reese Kneeland,
  Tong Chen, Ashutosh Narang, Charan Santhirasegaran, Jonathan Xu, Thomas
  Naselaris, Kenneth~A Norman, et~al.
\newblock Mindeye2: {S}hared-subject models enable {fMRI}-to-image with 1 hour
  of data.
\newblock In \emph{International Conference on Machine Learning},
  2024{\natexlab{b}}.

\bibitem[Shen et~al.(2019)Shen, Horikawa, Majima, and Kamitani]{shen2019deep}
Guohua Shen, Tomoyasu Horikawa, Kei Majima, and Yukiyasu Kamitani.
\newblock Deep image reconstruction from human brain activity.
\newblock \emph{PLoS Computational Biology}, 15\penalty0 (1):\penalty0
  e1006633, 2019.

\bibitem[Singh et~al.(2023)Singh, Pandey, Miyapuram, and
  Raman]{singh2023eeg2image}
Prajwal Singh, Pankaj Pandey, Krishna Miyapuram, and Shanmuganathan Raman.
\newblock {EEG2IMAGE}: image reconstruction from {EEG} brain signals.
\newblock In \emph{ICASSP 2023-2023 IEEE International Conference on Acoustics,
  Speech and Signal Processing}, pages 1--5. IEEE, 2023.

\bibitem[Song et~al.(2022)Song, Zheng, Liu, and Gao]{song2022eeg}
Yonghao Song, Qingqing Zheng, Bingchuan Liu, and Xiaorong Gao.
\newblock Eeg conformer: Convolutional transformer for eeg decoding and
  visualization.
\newblock \emph{IEEE Transactions on Neural Systems and Rehabilitation
  Engineering}, 31:\penalty0 710--719, 2022.

\bibitem[Song et~al.(2024)Song, Liu, Li, Shi, Wang, and Gao]{song2023decoding}
Yonghao Song, Bingchuan Liu, Xiang Li, Nanlin Shi, Yijun Wang, and Xiaorong
  Gao.
\newblock Decoding natural images from {EEG} for object recognition.
\newblock In \emph{The Twelfth International Conference on Learning
  Representations}, 2024.

\bibitem[Sun et~al.(2024{\natexlab{a}})Sun, Li, Chen, and
  Moens]{sun2024neurocine}
Jingyuan Sun, Mingxiao Li, Zijiao Chen, and Marie-Francine Moens.
\newblock Neurocine: Decoding vivid video sequences from human brain activties.
\newblock \emph{arXiv preprint arXiv:2402.01590}, 2024{\natexlab{a}}.

\bibitem[Sun et~al.(2024{\natexlab{b}})Sun, Li, Chen, Zhang, Wang, and
  Moens]{sun2024contrast}
Jingyuan Sun, Mingxiao Li, Zijiao Chen, Yunhao Zhang, Shaonan Wang, and
  Marie-Francine Moens.
\newblock Contrast, attend and diffuse to decode high-resolution images from
  brain activities.
\newblock \emph{Advances in Neural Information Processing Systems}, 36,
  2024{\natexlab{b}}.

\bibitem[Sur and Sinha(2009)]{sur2009event}
Shravani Sur and Vinod~Kumar Sinha.
\newblock Event-related potential: An overview.
\newblock \emph{Industrial Psychiatry Journal}, 18\penalty0 (1):\penalty0
  70--73, 2009.

\bibitem[Takagi and Nishimoto(2023)]{takagi2023high}
Yu Takagi and Shinji Nishimoto.
\newblock High-resolution image reconstruction with latent diffusion models
  from human brain activity.
\newblock In \emph{Proceedings of the IEEE/CVF Conference on Computer Vision
  and Pattern Recognition}, pages 14453--14463, 2023.

\bibitem[Vahdat et~al.(2022)Vahdat, Williams, Gojcic, Litany, Fidler, Kreis,
  et~al.]{vahdat2022lion}
Arash Vahdat, Francis Williams, Zan Gojcic, Or Litany, Sanja Fidler, Karsten
  Kreis, et~al.
\newblock Lion: {L}atent point diffusion models for {3D} shape generation.
\newblock \emph{Advances in Neural Information Processing Systems},
  35:\penalty0 10021--10039, 2022.

\bibitem[Vaswani et~al.(2017)Vaswani, Shazeer, Parmar, Uszkoreit, Jones, Gomez,
  Kaiser, and Polosukhin]{vaswani2017attention}
Ashish Vaswani, Noam Shazeer, Niki Parmar, Jakob Uszkoreit, Llion Jones,
  Aidan~N. Gomez, Lukasz Kaiser, and Illia Polosukhin.
\newblock Attention is all you need.
\newblock \emph{Advances in Neural Information Processing Systems}, 2017.

\bibitem[Wang et~al.(2022)Wang, Yan, Huang, Li, Wang, Fan, Sheng, Liu, Li, and
  Chen]{wang2022reconstructing}
Chong Wang, Hongmei Yan, Wei Huang, Jiyi Li, Yuting Wang, Yun-Shuang Fan, Wei
  Sheng, Tao Liu, Rong Li, and Huafu Chen.
\newblock Reconstructing rapid natural vision with {fMRI}-conditional video
  generative adversarial network.
\newblock \emph{Cerebral Cortex}, 32\penalty0 (20):\penalty0 4502--4511, 2022.

\bibitem[Wang and Isola(2020)]{wang2020understanding}
Tongzhou Wang and Phillip Isola.
\newblock Understanding contrastive representation learning through alignment
  and uniformity on the hypersphere.
\newblock In \emph{International Conference on Machine Learning}, pages
  9929--9939. PMLR, 2020.

\bibitem[Wen et~al.(2018)Wen, Shi, Zhang, Lu, Cao, and Liu]{wen2018neural}
Haiguang Wen, Junxing Shi, Yizhen Zhang, Kun-Han Lu, Jiayue Cao, and Zhongming
  Liu.
\newblock Neural encoding and decoding with deep learning for dynamic natural
  vision.
\newblock \emph{Cerebral Cortex}, 28\penalty0 (12):\penalty0 4136--4160, 2018.

\bibitem[Wu et~al.(2023)Wu, Wang, Feng, Xie, and Mian]{wu2023sketch}
Zijie Wu, Yaonan Wang, Mingtao Feng, He Xie, and Ajmal Mian.
\newblock Sketch and text guided diffusion model for colored point cloud
  generation.
\newblock In \emph{Proceedings of the IEEE/CVF International Conference on
  Computer Vision}, pages 8929--8939, 2023.

\bibitem[Xu et~al.(2024)Xu, Wang, Wang, Chen, Pang, and Lin]{xu2023pointllm}
Runsen Xu, Xiaolong Wang, Tai Wang, Yilun Chen, Jiangmiao Pang, and Dahua Lin.
\newblock Pointllm: {E}mpowering large language models to understand point
  clouds.
\newblock \emph{European Conference on Computer Vision}, 2024.

\bibitem[Yi et~al.(2024)Yi, Wang, Ren, and Li]{yi2024learning}
Ke Yi, Yansen Wang, Kan Ren, and Dongsheng Li.
\newblock Learning topology-agnostic eeg representations with geometry-aware
  modeling.
\newblock \emph{Advances in Neural Information Processing Systems}, 36, 2024.

\bibitem[Yu et~al.(2024)Yu, Li, Fu, Miao, and Cui]{yu2024accelerating}
Zihao Yu, Haoyang Li, Fangcheng Fu, Xupeng Miao, and Bin Cui.
\newblock Accelerating text-to-image editing via cache-enabled sparse diffusion
  inference.
\newblock In \emph{Proceedings of the AAAI Conference on Artificial
  Intelligence}, pages 16605--16613, 2024.

\bibitem[Zhang et~al.(2023)Zhang, Rao, and Agrawala]{zhang2023adding}
Lvmin Zhang, Anyi Rao, and Maneesh Agrawala.
\newblock Adding conditional control to text-to-image diffusion models.
\newblock In \emph{Proceedings of the IEEE/CVF International Conference on
  Computer Vision}, pages 3836--3847, 2023.

\bibitem[Zhang et~al.(2022)Zhang, Yu, Liu, and Huang]{zhang2022neural}
Yi-Jun Zhang, Zhao-Fei Yu, Jian~K Liu, and Tie-Jun Huang.
\newblock Neural decoding of visual information across different neural
  recording modalities and approaches.
\newblock \emph{Machine Intelligence Research}, 19\penalty0 (5):\penalty0
  350--365, 2022.

\bibitem[Zhou et~al.(2021)Zhou, Du, and Wu]{zhou20213d}
Linqi Zhou, Yilun Du, and Jiajun Wu.
\newblock {3D} shape generation and completion through point-voxel diffusion.
\newblock In \emph{Proceedings of the IEEE/CVF International Conference on
  Computer Vision}, pages 5826--5835, 2021.

\bibitem[Zhu et~al.(2023)Zhu, Chen, Shen, Li, and Elhoseiny]{zhu2023minigpt}
Deyao Zhu, Jun Chen, Xiaoqian Shen, Xiang Li, and Mohamed Elhoseiny.
\newblock Minigpt-4: Enhancing vision-language understanding with advanced
  large language models.
\newblock \emph{arXiv preprint arXiv:2304.10592}, 2023.

\end{thebibliography}
	}
	
\clearpage
\setcounter{page}{1}
\maketitlesupplementary

\section{EEG Data Preprocessing}

In this section, we introduce the details of EEG preprocessing pipeline. During data acquisition, static 3D image and dynamic 3D video stimuli were preceded by a marker to streamline subsequent data processing. 
The continuous EEG recordings were subsequently preprocessed using MNE \cite{gramfort2013meg}. The data were segmented into fixed-length epochs (1s for static stimuli and 6s for dynamic stimuli), time-locked to stimulus onset, with baseline correction achieved by subtracting the mean signal amplitude during the pre-stimulus period. The signals were downsampled from 1000 Hz to 250 Hz, and a bandpass filter (0.1–100 Hz) was applied in conjunction with a 50 Hz notch filter to mitigate noise. To normalize signal amplitude variability across channels, multivariate noise normalization was employed \cite{guggenmos2018multivariate}. Consistent with established practices \cite{li2024visual}, two stimulus repetitions were treated as independent samples during training to enhance learning, while testing involved averaging across four repetitions to improve the signal-to-noise ratio, following principles similar to those used in Event-Related Potential (ERP) analysis \cite{sur2009event}.

\section{Evaluation Metrics for Reconstruction Benchmark}

To assess the quality of the generated outputs, we adopt the N-way, top-K metric, a standard approach in 2D image decoding \cite{li2024visual,chen2024cinematic,chen2023seeing}. For 2D image evaluation, a pre-trained ImageNet1K classifier is used to classify both the generated images and their corresponding ground truth images. Similarly, we utilize data from Objaverse \cite{deitke2023objaverse} to pre-train a PointNet++ model \cite{qi2017pointnet++}. To ensure classifier reliability, the network is trained on all Objaverse data with category labels, excluding the test set used in our study. The point cloud data corresponding to the 3D objects is sourced from \cite{xu2023pointllm}. During evaluation, both the generated point clouds and their corresponding ground truth point clouds are classified using the trained network. The results are then analyzed to confirm whether the reconstructed object is correctly identified within the top K categories among N selected.
For the efficiency of evaluation, we utilize data from the first five subjects to train and evaluate the reconstruction model.
Moreover, a distinct feature of the diffusion model is its dependence on initialization noise, which can influence the generated outputs. We perform five independent inferences for each object and compute the average N-way, top-K metric across these runs. Additionally, to capture the potential best-case performance, we identify the optimal result based on the classifier's predicted scores across the five inferences and compute the N-way, top-K metric. 

\section{Analysis of Individual Difference}

\begin{figure}[t]
  \centering
  \begin{subfigure}{0.9\linewidth}
    \centering
    \includegraphics[width=\linewidth]{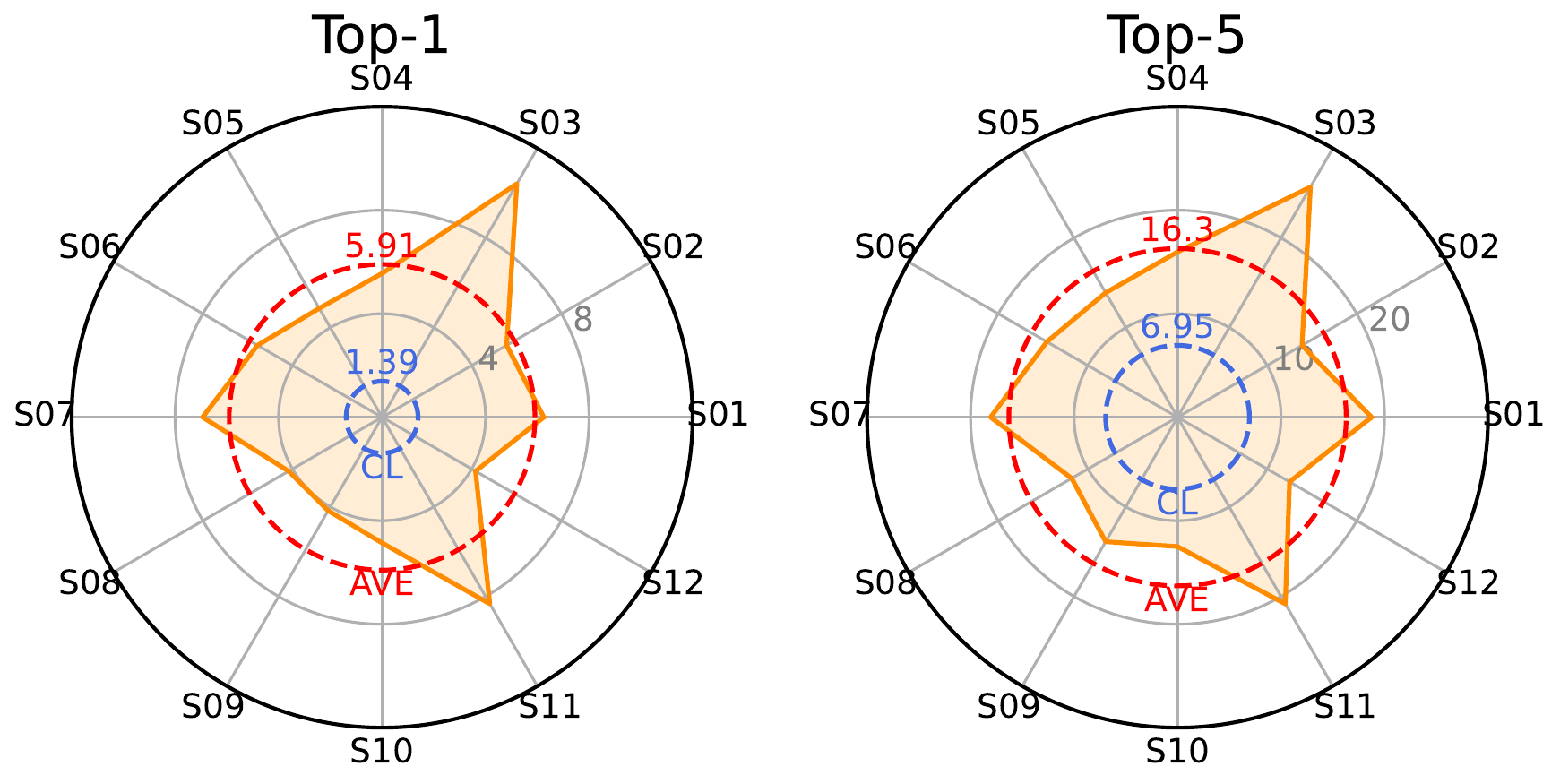} 
    \caption{Object Classification}
    \label{fig_areas1}
  \end{subfigure}

  \begin{subfigure}{0.85\linewidth}
    \centering
    \includegraphics[width=\linewidth]{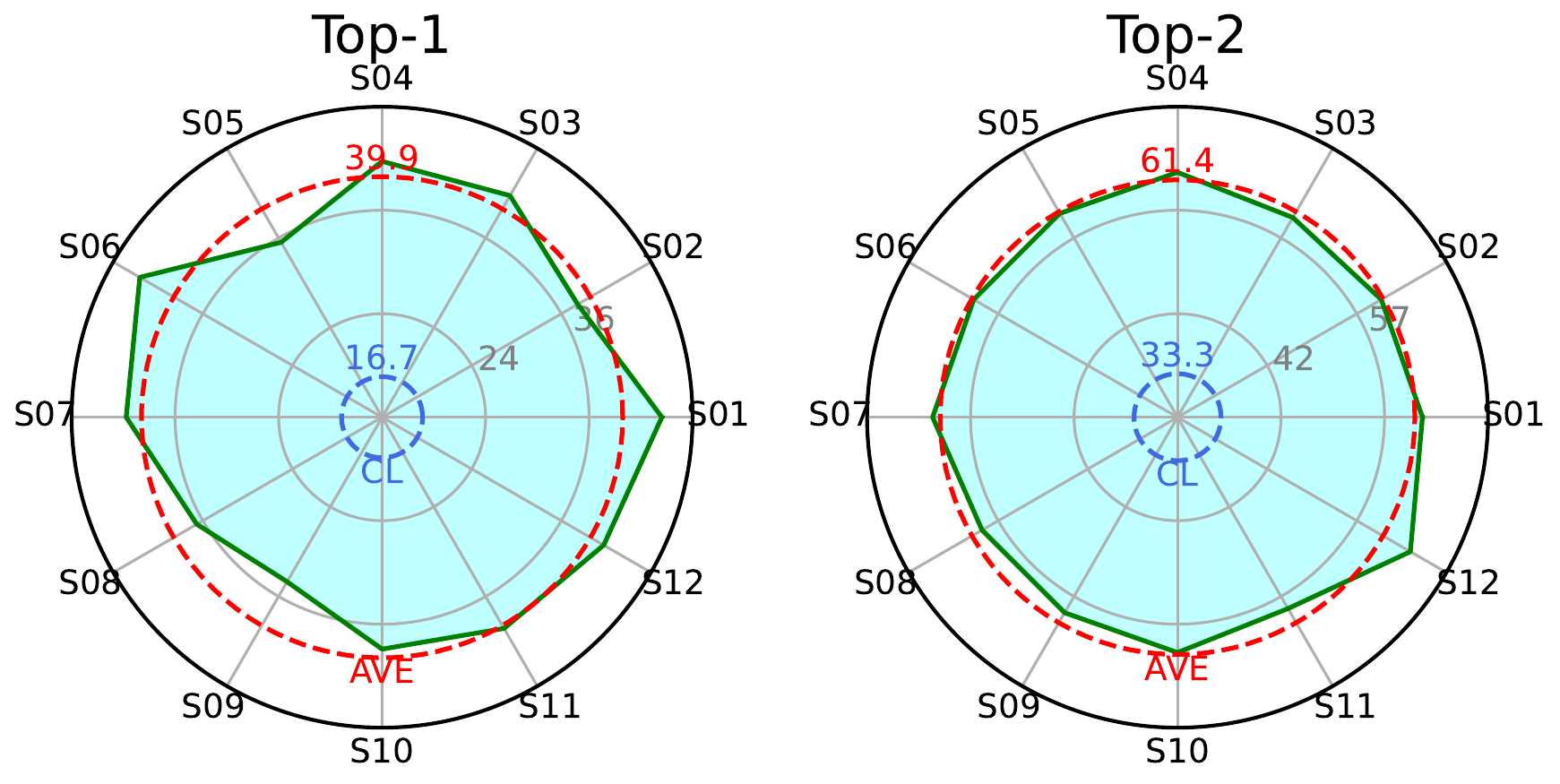}
    \caption{Color Classification}
    \label{fig_areas2}
  \end{subfigure}
  \caption{The results of individual analysis. (a) presents the top-1 and top-5 accuracies for the object classification task across 12 subjects, while (b) depicts the top-1 and top-2 accuracies for the classification task. The \textcolor{blue}{blue} line in each panel indicates chance-level performance, and the \textcolor{red}{red} line represents the average performance across all subjects. }
  \label{fig_individual}
  \vspace{-0.4cm}
\end{figure}

\begin{figure*}[h]
    \centering
    \includegraphics[width=0.9\textwidth]{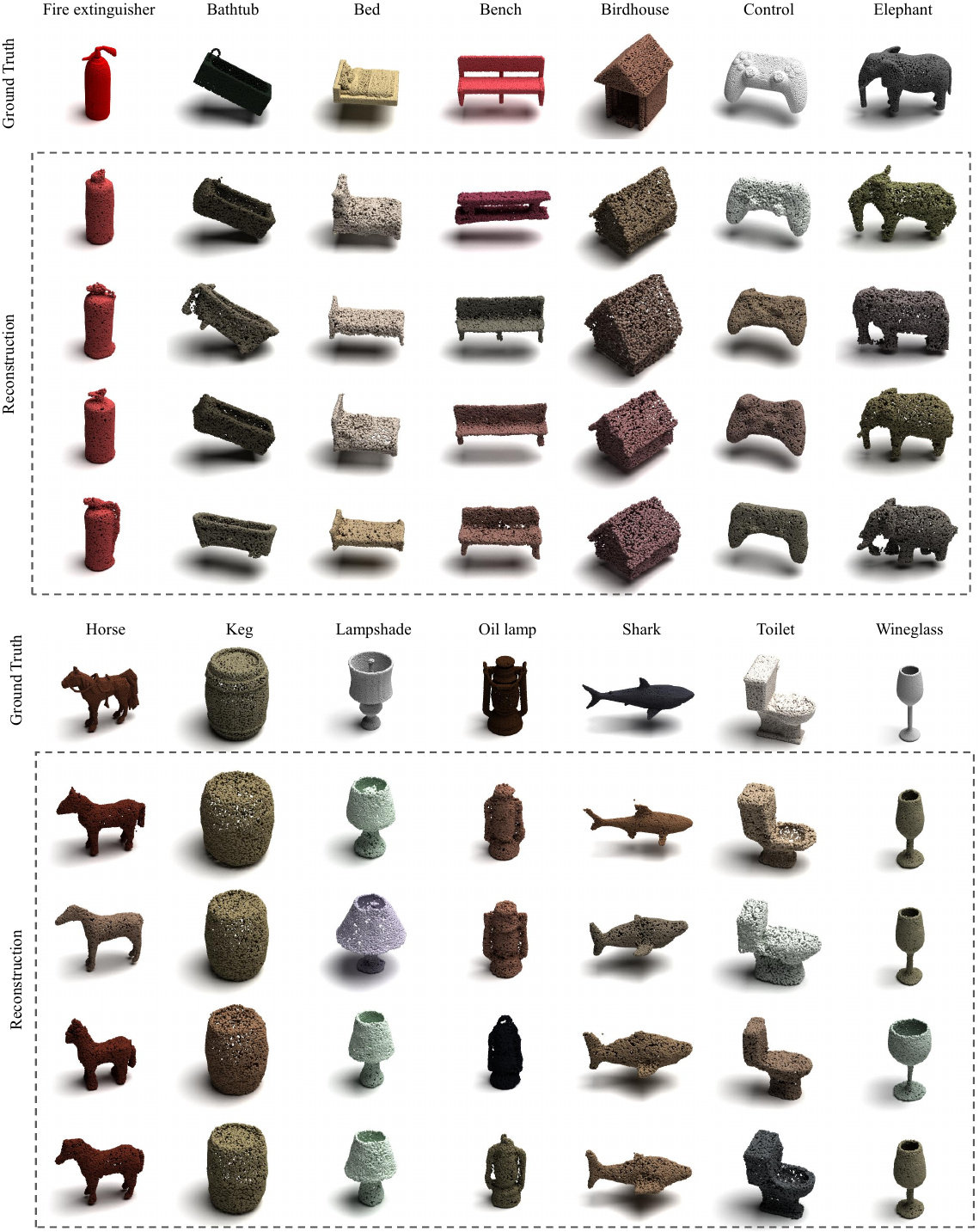} 
    \caption{More results reconstructed by Neuro-3D with different samplings trials, and the corresponding ground truth. The sampling variations arise either from results obtained across different subjects or from inference outputs of the diffusion model for the same subject using distinct noise initializations.}
    \label{fig-more}
\end{figure*}

We present the performance variability across individuals on two classification tasks, as illustrated in Fig. \ref{fig_individual}. On both tasks, individual performance consistently exceeds chance level, demonstrating that EEG signals encode visual perception information and that our method effectively extracts and utilizes this information for decoding. Notably, performance varies across tasks for the same individual. For instance, participant $S12$ performs significantly below average in object classification but achieves above-average results in color classification, suggesting distinct neural mechanisms underlying the processing of different visual attributes and their representation in EEG signals.

Furthermore, it has been widely confirmed that EEG signal has substantial individual variations \cite{gibson2022eeg,saha2020intra,huang2023discrepancy}. As shown in Fig. \ref{fig_individual}, significant differences are observed between individuals performing the same task, particularly in object classification, where $S03$ and $S11$ exhibit superior performance, while $S08$, $S09$ and $S12$ fall markedly below average. Similar variability is observed in the color classification, albeit to a lesser extent. These results verify the pronounced inter-subject differences in EEG signals and highlight a critical challenge for cross-subject EEG visual decoding, where performance remains suboptimal. Addressing this variability is a key focus for future research.
\vspace{-0.1cm}
\section{More Reconstructed Samples}
\vspace{-0.1cm}
Additional reconstructed results alongside their corresponding ground truth point clouds are presented in Fig. \ref{fig-more}. The proposed Neuro-3D framework exhibits robust performance, effectively capturing semantic categories, shape details, and the overall color of various objects.
\vspace{-0.1cm}
\section{Analysis of Failure Cases}
\begin{figure}[!h]
	\centering
	\includegraphics[width=1.0\columnwidth]{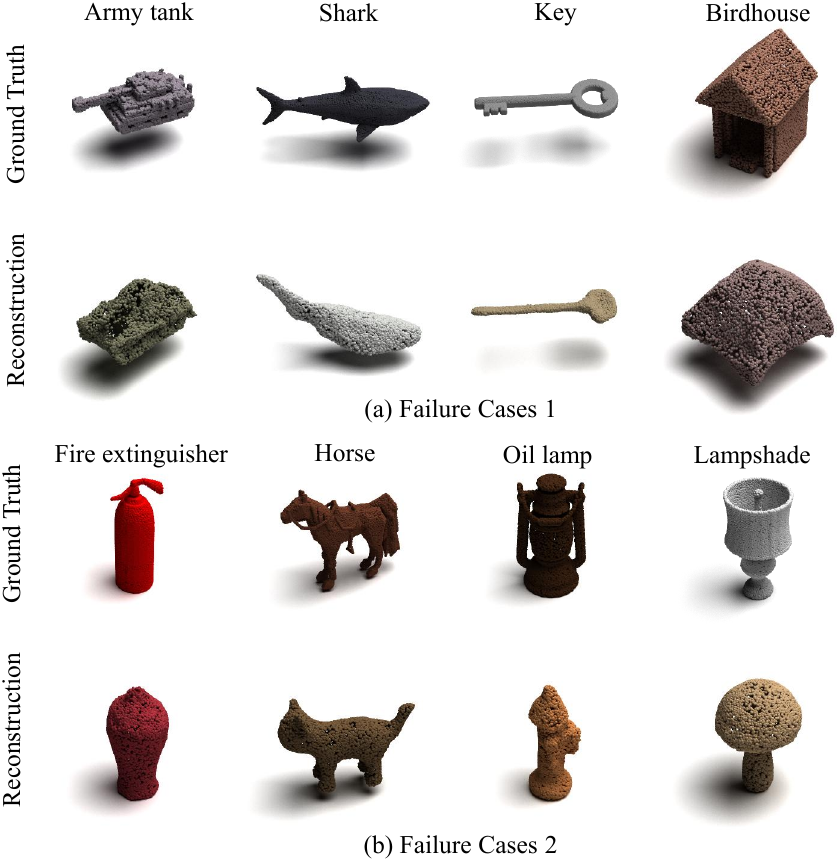} 
	\caption{Failure cases. (a) highlights reconstructions with significant loss of fine details, while (b) demonstrates several instances of incorrect semantic category prediction.}
	\label{fig-failure}
\end{figure}

\vspace{-0.1cm}
Fig. \ref{fig-failure} illustrates representative failure cases, categorized into two principal types: inaccuracies in detailed shape prediction and semantic reconstruction errors. Despite these limitations, certain features of the stimulus objects, including shape contours and color information, are partially preserved in the displayed reconstructed images. 
These shortcomings primarily arise from the inherent challenges of the low signal-to-noise ratio and limited spatial resolution of EEG signals, which constrain the performance of 3D object reconstruction. Addressing these issues presents a promising direction for future improvement.

\newpage
\vspace{0.2cm}

	
\end{document}